\documentclass{article}

\usepackage{PRIMEarxiv}

\usepackage[utf8]{inputenc} 
\usepackage[T1]{fontenc}    
\usepackage{hyperref}       
\usepackage{url}            
\usepackage{booktabs}       
\usepackage{amsfonts}       
\usepackage{nicefrac}       
\usepackage{microtype}      
\usepackage{lipsum}
\usepackage{fancyhdr}       
\usepackage{graphicx}       
\graphicspath{{media/}}     

\usepackage{amsmath,amssymb,amsfonts,bm,amsthm}
\usepackage{algorithmic}
\usepackage{graphicx}
\usepackage{textcomp}
\usepackage{xcolor}
\usepackage{booktabs}
\usepackage{multirow}
\usepackage{caption}
\usepackage{subcaption}
\usepackage{enumitem}
\usepackage{balance}
\usepackage{array}
\newtheorem{assumption}{Assumption}
\newtheorem{definition}{Definition}

\newcolumntype{P}[1]{>{\centering\arraybackslash}p{#1}} 
\newcolumntype{M}[1]{>{\centering\arraybackslash}m{#1}} 

\usepackage{comment}
\usepackage[linesnumbered,ruled,vlined]{algorithm2e}
\SetKwComment{Comment}{$\triangleright$\ }{}
\SetKwRepeat{Do}{do}{while}

\newcommand{\ours}{\textbf{CAPE}}

\newcommand{\bart}{\textbf{BART}}
\newcommand{\cf}{\textbf{CF}}
\newcommand{\cfrwass}{\textbf{$\text{CFR-WASS}$}}
\newcommand{\cfrmmd}{\textbf{$\text{CFR-MMD}$}}

\newcommand{\cevae}{\textbf{CEVAE}}
\newcommand{\netdeconf}{\textbf{Net-Deconf}}

\newcommand{\site}{\textbf{SITE}}

\newcommand{\cola}{\textbf{Cola-GNN}}
\newcommand{\gwnet}{\textbf{GWNet}}

\title{Causal Knowledge Guided Societal Event Forecasting}

\author{
  Songgaojun Deng \\
  Stevens Institute of Technology \\
  Hoboken, NJ\\
  \texttt{sdeng4@stevens.edu} \\
   \And
  Huzefa Rangwala \\
  George Mason University \\
  Fairfax, VA\\
  \texttt{rangwala@gmu.edu} \\
    \And
    Yue Ning \\
  Stevens Institute of Technology \\
  Hoboken, NJ\\
  \texttt{yue.ning@stevens.edu} \\
}

\begin{document}
\maketitle

\begin{abstract}
Data-driven societal event forecasting methods exploit relevant historical information to predict future events.
These methods rely on historical labeled data and cannot accurately predict events when data are limited or of poor quality.
Studying causal effects between events goes beyond correlation analysis and can contribute to a more robust prediction of events.
However, incorporating causality analysis in data-driven event forecasting is challenging due to several factors: (i) Events occur in a complex and dynamic social environment. Many unobserved variables, i.e., hidden confounders, affect both potential causes and outcomes. (ii) Given spatiotemporal non-independent and identically distributed (non-IID) data, modeling hidden confounders for accurate causal effect estimation is not trivial. 
In this work, we introduce a deep learning framework that integrates causal effect estimation into event forecasting.
We first study the problem of Individual Treatment Effect (ITE) estimation from observational event data with spatiotemporal attributes and present a novel causal inference model to estimate ITEs. 
We then incorporate the learned event-related causal information into event prediction as prior knowledge.
Two robust learning modules, including a feature reweighting module and an approximate constraint loss, are introduced to enable prior knowledge injection.
We evaluate the proposed causal inference model on real-world event datasets and validate the effectiveness of proposed robust learning modules in event prediction by feeding learned causal information into different deep learning methods.
Experimental results demonstrate the strengths of the proposed causal inference model for ITE estimation in societal events and showcase the beneficial properties of robust learning modules in societal event forecasting.
\end{abstract}

\keywords{Event Forecasting \and Representation Learning  \and Causal Inference}

\section{Introduction}
Predicting large-scale societal events, such as disease outbreaks, organized crime, and civil unrest movements, from social media streams, web logs, and news media is of great significance for decision-making and resource allocation. 
Previous methods mainly focus on improving the predictive accuracy of a given event type or multiple event types using historical event data~\cite{zhao2015spatiotemporal,zhao2016hierarchical}. Recently, to enhance model explainability, many approaches identify salient features or supporting evidence, such as precursor documents~\cite{ning2016modeling}, relationships represented as graphs~\cite{deng2019learning}, and major actors participating in the events~\cite{10.1145/3394486.3403209}. 
However, existing work explains the occurrence of events based on correlation-based indicators.

Attempts to study causality in event analysis and prediction have focused on extracting pairs of causal events from unstructured text ~\cite{radinsky2012learning}, or using human-defined causally related historical events to predict events of interest~\cite{radinsky2013mining}.
Causal effect learning has shown advantages in improving predictions in various machine learning problems, such as recommender systems~\cite{bonner2018causal}, disease diagnosis prediction~\cite{li2020teaching}, and computer vision tasks~\cite{chen2021spatial}.
This suggests the potential of causal effect learning for better prediction of societal events.
Leveraging causal effects can presumably provide new insights into causal-level interpretation and improve the robustness of event prediction, e.g., less susceptible to noise in data.
In this study, we explore societal event forecasting methods with the help of causal effect learning.

Traditionally, learning causal effects (aka treatment effects) from observational data involves estimating causal effects of a treatment variable (e.g., medication) on an outcome variable (e.g., recovery) given observable covariates (e.g., gender). 
In practice, there are also unobserved covariates, i.e., \textit{hidden confounders}, that affect both treatment and outcome variables. 
For instance, consider a study to evaluate the effectiveness of a medication. Gender as a covariate affects whether a patient chooses to take the medication and the corresponding outcome. The patient's living habits can be hidden confounders that affect both the patient's medication and outcome.
Exploring hidden confounders allows for more accurate estimations of treatment effects~\cite{louizos2017causal,guo2019learning,10.1145/3437963.3441818}. 

In this work, we formulate the problem of estimating treatment effects in the context of societal events.
Societal events can be classified into different types.
Given a time window, we look at multiple types of events (e.g., ``appeal'', ``investigation'') at a location and define treatment variables to be the detection of increased counts of these events compared to the previous time window. 
If the sudden and frequent occurrence of such events triggers some event of interest, the implied causal effect can be used to guide and interpret event predictions.
We define the outcome as the occurrence of an event of interest (e.g., ``protest'') at a future time. 
Both treatment and outcome variables can be affected by hidden social factors (i.e., hidden confounders) that are difficult to explicitly capture due to complex dependencies.
Intuitively, exploring hidden confounders can allow us to estimate causal effects more accurately.
To this end, we formulate our main research question as: \textit{can we build a robust event predictive model by incorporating treatment effect estimation with hidden confounder learning?} 
There are some challenges in solving this problem:
\begin{itemize}
\item 
Societal events have geographical characteristics and exhibit a high degree of temporal dependency~\cite{zhao2016hierarchical,ning2016modeling,10.1145/3394486.3403209}. 
Modeling spatiotemporal information requires an in-depth investigation of the dynamic spatial dependencies of societal events. 
However, few studies have focused on modeling spatiotemporal dependencies in causal effect learning, which poses a challenge for learning causal effects from societal events.
\item 
Events occur in a complex and evolving social environment. Many unknown social factors increase the difficulty of accurately estimating causal effects of events. 
Moreover, events are often caused by a variety of factors rather than a single determinant.
Utilizing causal effects to assist in event prediction is a new challenge.
\end{itemize}

We address the above challenges by first introducing the task of Individual Treatment Effect (ITE) estimation from societal events.
ITE is defined as the expected difference between the treated outcome and control outcome, where the outcome is the occurrence of a future event (e.g., \textit{protest}) at a specific place and time, and the treatment is a change in some event (e.g., \textit{appeal}) in the past.
We consider multiple treatments (e.g., \textit{appeal}, \textit{investigation}, etc.) with the motivation that the underlying causes of societal events are often complex. 
We model the spatiotemporal dependencies in learning the representations of hidden confounders to estimate ITEs. 
We then present an approach to inject the learned causal information into a data-driven predictive model to improve its predictive power.
Our contributions are summarized as follows:
\begin{itemize}
 \item We introduce a novel causal inference model for ITE estimation, which learns the representation of hidden confounders by capturing spatiotemporal dependencies of events in different locations.
 \item We propose two robust learning modules for event prediction that take as prior knowledge the information learned from the causal inference model. Incorporating such modules can enable event prediction models to be more robust to data noise.
\end{itemize}
We evaluate the proposed method against other state-of-the-art methods on several real-world event datasets. Through extensive experiments, we demonstrate the strengths of the proposed method in treatment effect learning and robust event prediction.

\section{Related Work}
\subsection{Event Prediction}
Event prediction focuses on forecasting future events that have not yet happened based on various social indicators, such as event occurrence rates and news reports.
Related research has been conducted in various fields and applications, such as election prediction~\cite{tumasjan2010predicting,o2010tweets}, stock market forecasting~\cite{bollen2011twitter}, disease outbreak simulation~\cite{signorini2011use,achrekar2011predicting}, and crime prediction~\cite{wang2012automatic}. 
Machine learning models such as linear regression~\cite{bollen2011twitter} and random forests~\cite{kallus2014predicting} were investigated to predict events of interest.
Time-series methods such as autoregressive models were studied to capture the temporal evolution of event-related indicators~\cite{achrekar2011predicting}.
With the increased availability of various data, more sophisticated features have been shown effective in predicting societal events such as topic-related keywords~\cite{zhao2015spatiotemporal}, document embedding~\cite{ning2016modeling}, word graphs~\cite{deng2019learning} and knowledge graphs~\cite{10.1145/3394486.3403209,deng2021understanding}. More advanced machine learning and deep learning-based models have emerged, such as multi-instance learning~\cite{ning2016modeling}, multi-task learning~\cite{zhao2015multi,gao2019incomplete} and graph neural networks~\cite{deng2019learning,10.1145/3394486.3403209}. 
Given the spatiotemporal dependencies of events, some existing research work studied spatiotemporal correlations in event prediction~\cite{gerber2014predicting,wang2012spatio,zhao2014unsupervised}. 
However, few studies explored the causality in event prediction.
Our proposed model incorporates causal effect learning in a spatiotemporal event prediction framework. This gives us the benefit of discovering the effects of different potential causes on predicting future events.

\subsection{Individual Treatment Effect Estimation}
Individual treatment effect (ITE) estimation refers to estimating the causal effect of a treatment variable on its outcome.
A wealth of observational data facilitates treatment effect estimation in many fields, such as health care~\cite{anglemyer2014healthcare}, education~\cite{gustafsson2013causal}, online advertising~\cite{sun2015causal}, and recommender systems~\cite{bonner2018causal}.
Several methods have been studied for ITE estimation including regression and tree based model~\cite{hill2011bayesian,chipman2010bart}, counterfactual inference~\cite{johansson2016learning}, and representation learning~\cite{shalit2017estimating}. 
The former approaches rely on the Ignorability assumption~\cite{rosenbaum1983central}, which is often untenable in real-world studies. 
A deep latent variable model, CEVAE~\cite{louizos2017causal} learns representations of confounders through variational inference.
Recent work relaxed the Ignorability assumption and studied ITE estimation from observational data with an auxiliary network structure in a static~\cite{guo2019learning} or dynamic environment~\cite{10.1145/3437963.3441818}. 
In addition to the traditional causal effect estimation, a new study of causal inference, including multiple treatments and a single outcome, has emerged, namely, Multiple Causal Inference. Researchers have shown that compared with traditional causal inference, it requires weaker assumptions~\cite{wang2019blessings}.
ITE estimation would considerably benefit decision-making as it can provide potential outcomes with different treatment options.
Our work introduces ITE estimation to societal event studies and exploits event-related causal information for event forecasting.

\subsection{Knowledge Guided Machine Learning}
Purely data-driven approaches might lead to unsatisfactory results when limited data are available to train well-performing and sufficiently generalized models.
Such models may also break natural laws or other guidelines~\cite{von2019informed}.
These problems have led to an increasing amount of research that focuses on incorporating additional prior knowledge into the learning process to improve machine learning models.
For example, logic rules~\cite{diligenti2017integrating,xu2018semantic} or algebraic equations~\cite{karpatne2017physics,stewart2017label,muralidhar2018incorporating} have been added as constraints to loss functions.
Knowledge graphs are adopted to enhance neural networks with information about relations between instances~\cite{battaglia2016interaction}.
The growth of this research suggests that the combination of data and knowledge-driven approaches is becoming relevant and showing benefits in a growing number of areas.
Existing work has typically focused on pre-existing knowledge obtained by human experts.
However, such approaches fail when prior knowledge is not available, e.g., for societal events.
Some researchers explored causal knowledge-guided methods in health prediction~\cite{li2020teaching} and  image-to-video adaptation~\cite{chen2021spatial}.
In this work, we study causal effects between societal events and use the learned causal information as prior knowledge for event prediction.

\begin{table}[]
\small
    \centering
 \caption{Important notations and descriptions.}
    \label{tab:notation}
    \begin{tabular}{c| l}
 \toprule
    \textbf{Notations} &  \textbf{Descriptions} \\
 \midrule
 $M,T,E$ & sets of locations, timestamps and numbers of event types \\
 $\mathbf{X}_i^{\leq t }  \in \mathbb{R}^{E\times \Delta}$ & covariates of the $i$-th location before time $t$\\
 $\mathbf{x}_i^t \in \mathbb{R}^E$ & frequency of events at time $ t $ for location $ i $  \\ 
 $\mathbf{c}_i^{\leq t} \in \{0,1\}^{E}$ & observed treatment vector of the $i$-th location before time $t$ \\
 $c_{i(j)}^{\leq t} \in \{0,1\}$ & observed  $j$-th treatment of the $i$-th location before time $t$\\
$y^{t+\delta}_{i(j)}(1),y^{t+\delta}_{i(j)}(0)$ & potential outcomes for the $j$-th treatment of the $i$-th location at time $t+\delta$\\
$\hat{y}^{t+\delta}_{i(j)}(1),\hat{y}^{t+\delta}_{i(j)}(0)$ & predicted potential outcomes for the $j$-th treatment of the $i$-th location at time $t+\delta$\\
 $\mathbf{A} \in \mathbb{R}^{M\times M}$ & connectivity of $M$ locations \\
  $\tau_{i(j)}^{t+\delta}$ & ITE of the $i$-th location at time $t+\delta$ for the $j$-th treatment\\
 $\mathbf{z}_{i(j)}^{\leq t}$ & learned hidden confounders of the $i$-th location before time $t$ when the $j$-th treatment is considered \\
     \bottomrule
    \end{tabular}
\end{table}

\section{Problem Formulation}
The objective of this study is two-fold: (1) given multiple pre-defined treatment events (e.g., appeal, investigation, etc.), estimate their causal effect on a target event (i.e., protest) individually; (2) predict the probability of the target event occurring in the future with the help of estimated causal information.
In the following, we will introduce the observational event data, individual treatment effect learning, and event prediction.

\subsection{Observational Event Data} 
In this work, we focus on modeling the occurrence of one type of societal event (i.e., ``protest'') by exploring the possible effects it might receive from other types of events (e.g., ``appeals'' and ``investigation'').
A total of $E$ categories of societal events are studied.
These events happen at different locations and times.
We use $M, T$ to denote the sets of locations and timestamps of interest, respectively.
The observational event data can be denoted as 
$\mathcal{D} = \big \{  \{\mathbf{X}^{\leq t}_i,\mathbf{c}^{\leq t}_i, y^{t+\delta}_i, \}^{M}_{i=1}, \mathbf{A} \big \}_{t \in T}$,
where $\mathbf{X}^{\leq t}_i,\mathbf{c}^{\leq t}_i, y^{t+\delta}_i$ denote the pre-treatment covariates/features, observed treatments, and outcome, respectively. 
$\mathbf{A} \in \mathbb{R}^{M\times M}$ represents the connectivity of $M$ locations, where each element can denote a fixed geographic distance or the degree of influence of events between locations.
Important notations are presented in Table~\ref{tab:notation}.

\textbf{Covariates}:
We define the covariates $\mathbf{X}^{\leq t}_i = (\mathbf{x}^{t-\Delta+1}_i,...,\mathbf{x}^{t}_i)  \in \mathbb{R}^{E\times \Delta}$  to be the historical events at location $i$ with size $\Delta$ up to time $t$. 
$\mathbf{x}_i^t \in \mathbb{R}^E$ is a vector representing the frequencies of $E$ types of events that occurred at location $i$ at time $t$.

\textbf{Treatments}:
The treatments $\mathbf{c}_i^{\leq t} \in \{0,1\}^{E}$ can be represented by a binary vector with dimension $E$ where each element indicates the occurrence states of a type of events (e.g., appeal). 
Specifically, the $j$-th element $c_{i(j)}^{\leq t} = 1$ indicates a notable (i.e., 50\%) increase of the $j$-th event type at window $[t-\Delta+1,t]$ from the previous window $[t-2\Delta+1,t-\Delta]$.
\footnote{The comparison of two windows is motivated by studies showing that short-term historical data can lead to favorable performance in event prediction~\cite{10.1145/3394486.3403209,jin2020Renet}. A threshold of 50\% is selected heuristically.
We leave variant treatment settings for future work.}
A value of 1 means getting treated and 0 means getting controlled.
For convenience, we refer to each element in the treatment vector as a treatment event.\footnote{Our setup differs from multiple causal inference ~\cite{wang2019blessings,bica2020time},  which estimates the potential outcome of a combination of multiple treatments. We are more interested in studying the potential outcome of each element in the treatment vector.}

\textbf{Observed Outcome}:
The observed/factual outcome $y^{t+\delta}_i \in \{0,1\}$ is a binary variable denoting if an event of interest (i.e., protest) occurs at location $i$ in the future ($t+\delta$).
$\delta \geq 1$ is the lead time indicating the number of timestamps in advance for a prediction. 

\begin{figure}
    \centering 
    \includegraphics[width=.3\linewidth]{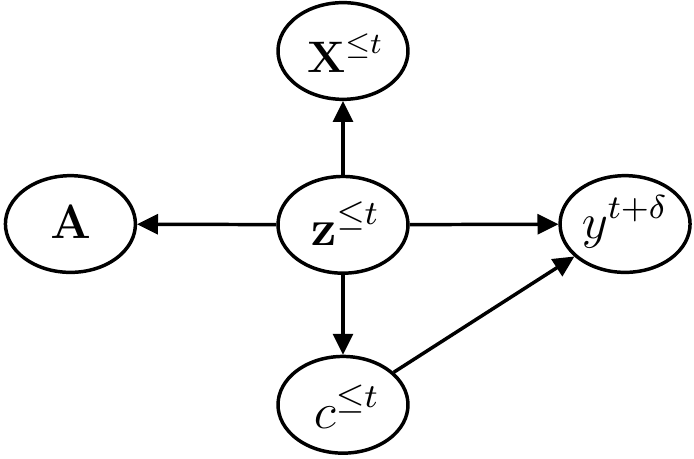}
    \caption{Causal graph defined for the ITE estimation on observational event data. The edges represent causal relations.
    For a location, we use $\mathbf{X}^{\leq t}, {c}^{\leq t}, \mathbf{z}^{\leq t},y^{t+\delta}$ to denote the covariates, assignment for a treatment event, representation of hidden confounders before time $t$, and outcome at time $t+\delta$, respectively. 
    $\mathbf{A}$ denotes the connectivity among locations.
    } 
    \label{fig:causal-graph}
\end{figure}

\subsection{Individual Treatment Effects Learning}
We first define potential outcomes in observational event data following well-studied causal inference frameworks~\cite{rubin1978bayesian,rubin2005causal}. We ignore the location subscript $i$ for simplicity unless otherwise stated. 

\textbf{Potential Outcomes}: In general, the potential outcome $Y(C)$ denotes what the outcome an instance would receive, if the instance was to take treatment $C$. A potential outcome $Y(C)$ is distinct from the observed/factual outcome $Y$ in that not all potential outcomes are observed in the real world. 
In our problem, there are two potential outcomes for each treatment event.
Given a location at time $t+\delta$ and the $j$-th treatment event, 
we denote by $y^{t+\delta}_{(j)}(1)$ the potential outcome (i.e., occurrence of protest) if the $j$-th treatment event is getting treated, i.e., $c_{(j)}^{\leq t}=1$. Similarly, we denote by $y^{t+\delta}_{(j)}(0)$ the potential outcome we would observe if the treatment event is under control, i.e., $c_{(j)}^{\leq t}=0$. 

The factual outcome is $y^{t+\delta}$ when the location has already received the treatment assignment before time $t$. 
The counterfactual outcome is defined if the location obtains the opposite treatment assignment. In the observational study, only the factual outcomes are available, while the counterfactual outcomes can never be observed.

\begin{figure}
    \centering 
    \includegraphics[width=0.71\linewidth]{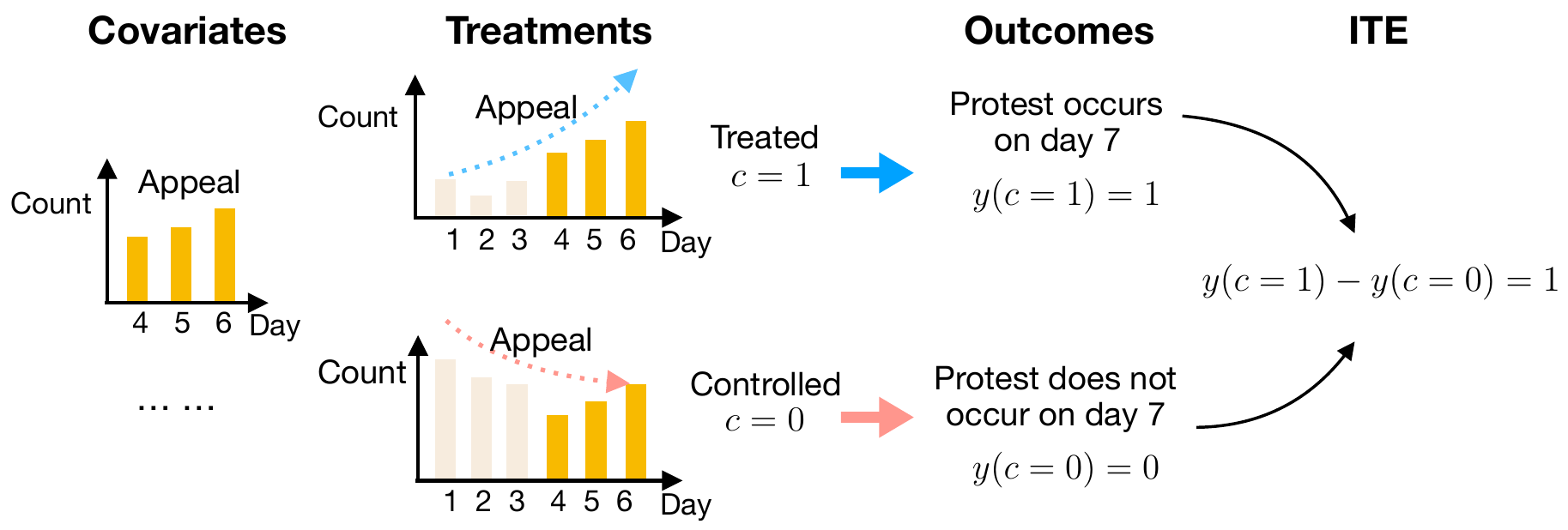}
    \caption{
    An example of ITE estimation on observational event data. 
    We use the horizontal axis to indicate time and the bars to indicate the count of events in a location. Taking the Appeal event as the treatment event, the treated and controlled scenarios denote whether the amount of appeal events in the current window (on average) is greater or less than the previous window. The window size is 3. The light orange bars indicate the appeal events of the past window. The outcomes denote whether the protest will occur at a future time given each treatment assignment. ITE is estimated from the two potential outcomes. 
    } 
    \label{fig:ite-example}
\end{figure}

The Individual Treatment Effect (ITE) is the difference between two potential outcomes of an instance, examining whether the treatment affects the outcome of the instance.
In observational event data, for the $j$-th treatment event, we formulate the ITE for the location at time $t+\delta$ in the form of the Conditional Average Treatment Effect (CATE)~\cite{shalit2017estimating,10.1145/3437963.3441818}:
\begin{equation}
    \tau_{(j)}^{t+\delta} = \mathbb{E}[y^{t+\delta}_{(j)}(1)-y^{t+\delta}_{(j)}(0)|\mathbf{X}^{\leq t},\mathbf{A}].
\end{equation}
We provide a toy example to illustrate ITE estimation on observational event data, as shown in Fig.~\ref{fig:ite-example}.

In this study, we aim to estimate ITEs and then use them for event prediction.
The challenge of ITE estimation lies in how to estimate the missing counterfactual outcome.
Our estimation of ITE is built upon some essential assumptions.
For simplicity and readability, we omit the subscripts for the location ${i}$ and the treatment event ${j}$ and use $c^{\leq t}$ to represent the $j$-th treatment event.

\begin{assumption}
\textbf{No Interference}. Assuming that one instance is defined as a location at a time in observational event data,
the potential outcome on one instance should be unaffected by the particular assignment of treatments on other instances. 
\end{assumption} 

\begin{assumption}
\textbf{Consistency}. The potential outcome of treatment $c^{\leq t}$ equals to the observed outcome if the actual treatment received $c^{\leq t}$, i.e., $ y^{t+\delta}(c^{\leq t}) = y^{t+\delta} $.
\end{assumption}

\begin{assumption}\textbf{Positivity}. 
If the probability $\mathbb{P}(\mathbf{X}^{\leq t},\mathbf{A}) \neq 0$, then the probability to receive treatment assignment 0 or 1 is positive, i.e., $0 < \mathbb{P}(c^{\leq t}=1|\mathbf{X}^{\leq t},\mathbf{A}) < 1$,
$c^{\leq t} \in \{0,1\}$.
\end{assumption}

The Positivity assumption indicates that before time $t$, each treatment assignment has a non-zero probability of being given to a location. This assumption is testable in practice.
In addition to these assumptions, most existing work~\cite{shalit2017estimating,louizos2017causal,wager2018estimation} relies on the Ignorability assumption, which assumes that all confounding variables are observed and reliably measured by a set of features for each instance, i.e., hidden confounders do not exist.

\begin{definition}
\textbf{Ignorability Assumption}. Given pre-treatment covariates $\mathbf{X}^{\leq t}$, the outcome variables are independent of its treatment assignment, $y^{t+\delta}(0),y^{t+\delta}(1)  \perp \!\!\! \perp c^{\leq t}|\mathbf{X}^{\leq t}$.
\end{definition}

However, this assumption is untenable in societal event studies due to the complex environment in which societal events occur.
We relax this assumption by introducing the existence of hidden confounders~\cite{guo2019learning}.
Note that hidden confounders are unobserved in observational event data but will be learned in our approach through a spatiotemporal model.
We define a causal graph, as shown in Fig.~\ref{fig:causal-graph}. 
The hidden confounders $\mathbf{z}^{\leq t}$ causally affect the treatment and outcome.\footnote{For the $j$-th treatment event, the hidden confounders can be written as $\mathbf{z}_{(j)}^{\leq t}$.} 
The potential outcomes are independent of the observed treatment, given the hidden confounders: $y^{t+\delta}(0),y^{t+\delta}(1)  \perp \!\!\! \perp c^{\leq t}|\mathbf{z}^{\leq t}$. 
In addition, we assume the features $\mathbf{X}^{\leq t}$  and the connectivity of locations $\mathbf{A}$ are proxy variables for hidden confounders $\mathbf{z}^{\leq t}$.  
Unobservable hidden confounders can be measured with $\mathbf{X}^{\leq t}$ and $\mathbf{A}$.
Based on the temporal and spatial characteristics of our observational event data. We introduce the following assumption~\cite{10.1145/3437963.3441818}: 

\begin{assumption}
\textbf{Spatiotemporal Dependencies in Hidden Confounders}.
In observational event data, hidden confounders capture spatial information among locations, reflected by $\mathbf{A}$, and show temporal dependencies of events across multiple historical steps (i.e., $\Delta$).
\end{assumption}

Note that this assumption does not contradict the \textit{No Interference} assumption. We focus on the scenario in which spatiotemporal information can be exploited to control confounding bias.

\subsection{Event Prediction} 
We present traditional event prediction and event prediction with causal knowledge proposed in this work.

\begin{definition}
\textbf{Event Prediction}. Learn a classifier that predicts the probability of the target event occurring at a location at time $t+\delta$ based on available data:
$    \mathbb{P}(y^{t+\delta}|\mathbf{X}^{\leq t}, \mathbf{A})$.
\end{definition}

Instead of learning a mapping function from input features to event labels, we are interested in estimating treatment effects under different treatment events individually and exploiting such causal information to enhance event prediction.

\begin{definition}
\textbf{Event Prediction with Causal Knowledge}.
Build an event forecaster using available data with causal information as prior knowledge: $\mathbb{P}(y^{t+\delta}|\mathbf{X}^{\leq t}, \mathbf{A}, \mathcal{C}(\mathbf{X}^{\leq t},\mathbf{c}^{\leq t}, \mathbf{A})  )$,
where $ \mathcal{C}$ is the trained causal inference model that takes features, multiple treatments and the connectivity information of locations as input and outputs potential outcomes.
\end{definition}
The multiple treatment setting (i.e., $\mathbf{c}^{\leq t}$) aims to produce informative causal knowledge to assist event prediction.  
We will discuss the proposed method of event prediction with causal knowledge in the following sections.

\section{Methodology}
\begin{figure*}
    \centering 
    \includegraphics[width=0.94\linewidth]{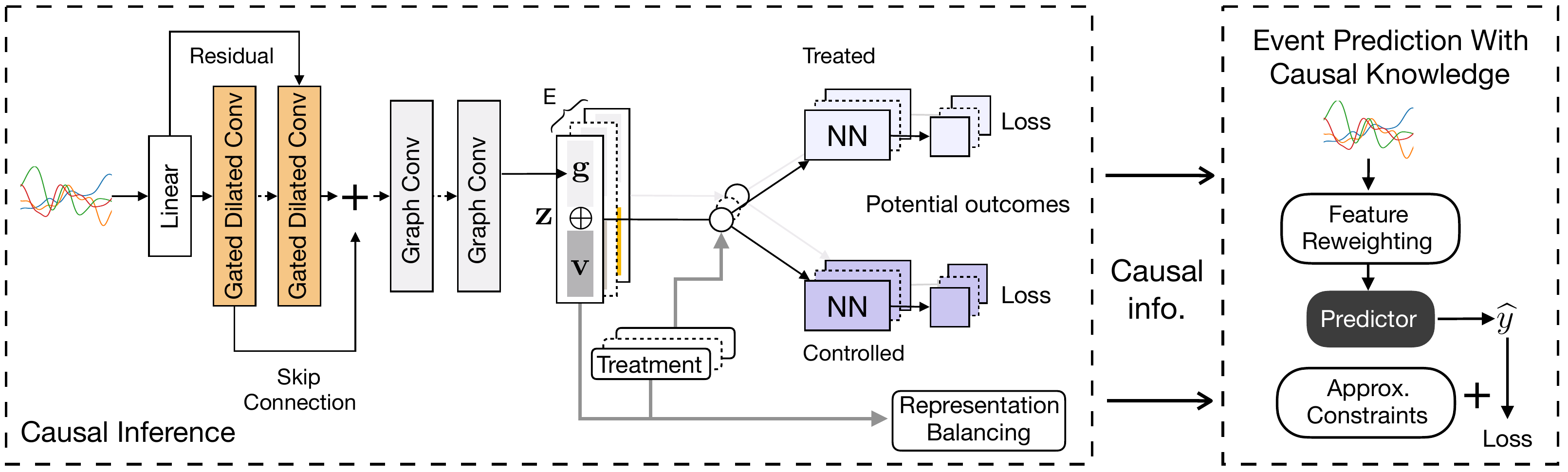} 
    \caption{
    The overall architecture of the proposed framework. 
    The proposed causal inference method learns to estimate ITEs in the presence of multiple treatment events of interest.
    The representation of hidden confounders is learned using a spatiotemporal model.
    Then, the vector representation of hidden confounders corresponding to each treatment event is fed into two neural networks (NNs) for individual treatment effect learning. 
    Next, the framework takes the output of the causal inference model as prior knowledge to forecast events.
    }
    \label{fig:framework}
\end{figure*}

We propose a novel framework \ours, which incorporates \textbf{ca}usal inference into the \textbf{p}rediction of future \textbf{e}vent occurrences in a spatiotemporal environment.\footnote{Code is available at \url{https://github.com/amy-deng/cape}}
In our framework, ITEs with different treatment events are jointly modeled in a spatiotemporal causal inference model. It will contribute to the final event prediction by feeding the causal output (e.g., potential outcomes) to a non-causal data-driven prediction model. 
The overall framework, as illustrated by Fig.~\ref{fig:framework}, consists of two parts: (1) causal inference and (2) event prediction. The causal inference component is designed to estimate the ITE, including two essential modules: hidden confounder learning and potential outcome prediction.
For each treatment event,
it learns the representation of hidden confounders by capturing spatiotemporal dependencies and outputs the potential outcomes under different treatment assignments.
The event prediction part comprises two robust learning modules, a feature reweighting module and an approximation constraint loss. They take the causal information learned from the causal inference model as prior knowledge to assist the training of a data-driven event prediction model.
Next, we will elaborate on these components.

\subsection{Causal Inference}
 \subsubsection{\textbf{Hidden Confounder Learning}}
Hidden confounders are common in real-world observational data~\cite{pearl2009causal}.
Assuming spatiotemporal dependencies exist in hidden confounders, we introduce a novel and effective network that models spatial and temporal information for each location at each time step. 
It consists of several temporal feature learning layers and spatial feature learning layers. 
Our network is based on the success of previous work~\cite{oord2016wavenet,dauphin2017language}.
It is designed to be adaptable to a multi-task setting to learn hidden confounders of multiple treatments.

\paragraph{Temporal Feature Learning} 
Dilated casual convolution networks~\cite{yu2015multi} handle long sequences in a non-recursive manner, which facilitates parallel computation and alleviates the gradient explosion problem. Gating mechanisms have shown benefits to control information flow through layers for convolution networks~\cite{oord2016wavenet,dauphin2017language,wu2019graph}.
We employ the dilated causal convolution with a gating mechanism in temporal feature learning to capture a location’s temporal dependencies. 
For a location before time $t$, the multivariate time series of historical event occurrences is a matrix $\mathbf{X}^{\leq t} \in \mathbb{R}^{E\times \Delta}$, where each row indicates the frequency sequence of one type of events in the historical window with size $\Delta$.  
We use a linear transformation to map the event frequency matrix into a latent space, i.e., $\mathbf{X}'^{\leq t} \in \mathbb{R}^{d_s\times \Delta}$. $d_s$ indicates the feature dimension in the latent space.
Then, we apply the dilated convolution to the sequence. For simplicity, we use $\mathbf{r} \in \mathbb{R}^{\Delta}$ to denote a row in the matrix $\mathbf{X}'^{\leq t}$.
Formally, for the 1-D sequence input $\mathbf{r}$ and a filter $\bm{\Gamma}  \in \mathbb{R}^K$, the dilated causal convolution
operation on element $s$ of the sequence is defined as:
\begin{equation}
   (\mathbf{r}*_{d}\bm{\Gamma} )(s) = \sum_{k=0}^{K-1}\bm{\Gamma} (k)\mathbf{r}(s-d \cdot k),  
\end{equation}
where $*_d$ is a $d$-dilated convolution. $K$ is the filter size. $(s),(k),(s-d\cdot k)$ are indices of vectors.
$(\mathbf{r}*_{d}\bm{\Gamma})$ is the output vector.

We further incorporate a gated dilated convolutional layer which consists of two parallel dilated convolution layers:
\begin{equation}
   \mathbf{h} =  \tanh  (\mathbf{r}*_{d}\bm{\Gamma}_{1})   \odot \sigma  (\mathbf{r}*_{d}\bm{\Gamma}_{2}),\label{equ:gatetcn}
\end{equation}
where $\bm{\Gamma}_{1},\bm{\Gamma}_{2}$ are filters for dilated convolutional layers, and $\odot$ is the Hadamard product. 
$\tanh(\cdot)$ is to regularize the features. $\sigma(\cdot)$ is the sigmoid function that determines the ratio of information passed to the next layer. 
Specifically, we stack multiple gated dilated convolutional layers (Eq.~\ref{equ:gatetcn}) with increasing dilation factors (e.g., $d = 1, 2, 4$).
Residual and skip connections are applied to avoid the vanishing gradient problem~\cite{oord2016wavenet,wu2019graph}.
To this end, the temporal dependencies are captured, and we use $\mathbf{H}^t \in \mathbb{R}^{M \times d_s}$ to denote the learned temporal features for $M$ locations at time $t$.

\paragraph{Spatial Feature Learning}
Graph convolution is a powerful operation to learn representations of nodes given the graph structure. To capture the spatial dependencies, we adopt the graph convolutional network (GCN)~\cite{kipf2016semi} to learn the spatial influence from locations by treating each location as a node in graph:
 \begin{equation}
 \mathbf{G}^t = \text{ReLU} ( \mathbf{A}'\mathbf{H}^{t}\mathbf{W}^{g}  ),
\label{eq:gcn2}
 \end{equation}
where $\mathbf{W}^g$ is the weight matrix for a GCN layer. $\mathbf{G}^t$ denotes the spatiotemporal feature matrix referring to all locations, where each row $\mathbf{g}^t$ captures the historical information of a specific location as well as the neighboring locations. $\mathbf{A}'$ is a learnable adjacency matrix.
The geographical adjacency matrix of locations usually cannot represent the connectivity of locations in the context of societal event forecasting. Therefore, we adopt the self-adaptive adjacency matrix~\cite{wu2019graph}, which does not require any prior knowledge and is learned through training.
We randomly initialize two node embedding matrices with learnable parameters $\mathbf{E}_1,\mathbf{E}_2 \in \mathbb{R}^{M \times d_a}$. The self-adaptive adjacency matrix is defined as:
\begin{equation}
    \mathbf{A}' = \text{Softmax}\big ( \text{ReLU} (\mathbf{E}_1 \mathbf{E}_2^{T}) \big ),
\end{equation}
where the ReLU activation function eliminates weak connections and the Softmax applies normalization.

\paragraph{Hidden Confounder Learning}
 
To learn the representation of hidden confounders, we utilize the spatiotemporal feature and a learnable embedding specific to each treatment event (i.e., $\mathbf{v}_{(j)},  1 \leq j \leq E$). It is worth pointing out that in the proposed framework, we include multiple treatment events and expect to estimate the ITE corresponding to each treatment event.  
The treatment-specific embedding aims to capture latent information of each treatment event and distinguish the hidden confounder representations learned for each treatment effect learning task.
Similar ideas of task embeddings are studied in prior work~\cite{vuorio2019multimodal}.
Given a location and a time $t$, the representation of hidden confounders for the $j$-th treatment is:
\begin{equation}
    \mathbf{z}^{\leq t}_{(j)} =\mathbf{g}^t \oplus \mathbf{v}_{(j)},
\end{equation}
where $\oplus$ is concatenation.

\subsubsection{\textbf{Potential Outcome Prediction}}
Using the above components, we obtain the representation of hidden confounders $\mathbf{z}_{(j)}^{\leq t}$.
Following the predefined causal graph in Fig.~\ref{fig:causal-graph}, the learned hidden confounders can be used to estimate potential outcomes. 
We use two networks that output two potential outcomes of the $j$-th treatment event, respectively:
\begin{equation}
\hat{y}^{t+\delta}_{(j)}(1) = \Phi_{(j)}^1 \big (\mathbf{z}^{\leq t}_{(j)}\big ), \quad \hat{y}^{t+\delta}_{(j)}(0) = \Phi_{(j)}^0 \big (\mathbf{z}^{\leq t}_{(j)} \big ),
\end{equation}
where $\hat{y}^{t+\delta}_{(j)}(1/0)$ denotes the inferred potential outcomes when the $j$-th treatment event is getting treated or controlled.
$\Phi_{(j)}^1(\cdot), \Phi_{(j)}^0(\cdot)$ are parameterized by deep neural networks with a sigmoid function at the last layer. The networks are trained end-to-end, and one can estimate the potential outcomes under multiple treatment events.

\subsubsection{\textbf{Objective Function}}
\paragraph{Potential Outcome Loss}
We use the binary cross-entropy loss as the objective factual loss for predicting potential outcomes. When only the $j$-th treatment event is considered (i.e., the general case for treatment effect learning~\cite{shalit2017estimating,guo2019learning,10.1145/3437963.3441818}), the factual loss is:
\begin{equation} \label{eq1}
\mathcal{L}_{(j)}^{\mbox{fact}} = -\sum_{t\in T}\sum_{i\in M}  y_i^{t+\delta} \log  \hat{y}_{i(j)}^{t+\delta} 
+ (1-y_i^{t+\delta})\log  \big (1-\hat{y}_{i(j)}^{t+\delta} \big) ,
\end{equation}
where $y_i^{t+\delta}$ is the observed outcome for location $i$ at time $t+\delta$.
$\hat{y}_{i(j)}^{t+\delta} = \hat{y}_{i(j)}^{t+\delta}(c^{\leq t}_{i(j)}) $ is the predicted outcome given the observed treatment $c^{\leq t}_{i(j)} \in \{0,1\}$.
Since our model predicts potential outcomes for multiple treatment events, we express the total factual loss as follows:
\begin{equation}
    \mathcal{L}^{\mbox{fact}} = \sum_{1 \leq j \leq E} \mathcal{L}_{(j)}^{\mbox{fact}}  + \eta \cdot  \Omega(\Theta) ,
\end{equation}
where $\Omega(\Theta)$ stands for the $\ell_2$-norm regularization for all training parameters and $\eta$ is the weight for scaling the regularization term.

\paragraph{Representation Balancing}
Studies have proved that balancing the representations of treated and control groups would help mitigate the confounding bias and minimize the upper bound of the outcome
inference error~\cite{johansson2016learning,shalit2017estimating}. Therefore, we incorporate a representation balancing layer to force the distributions of hidden confounders of treated and controlled groups to be similar. Specifically, we adopt the integral probability metric (IPM)~\cite{shalit2017estimating} to measure the difference between the distributions of the treated instances and the controlled instances in terms of their hidden confounder representations:
\begin{equation}
    \mathcal{L}^{\mbox{disc}} =  \alpha \cdot  \text{IPM}(\mathcal{Z}_1,\mathcal{Z}_0),
\end{equation}
where $\mathcal{Z}_1 = \{\mathbf{z}_{i(j)}^{\leq t}\}_{i,t,j: c_{i(j)}^{\leq t} = 1 }, \mathcal{Z}_0 = \{\mathbf{z}_{i(j)}^{\leq t}\}_{i,t,j: c_{i(j)}^{\leq t} = 0 }$ indicate the sets of hidden confounders for samples (in a batch) in the treated group and controlled group, respectively.
The IPM can be Wasserstein or Maximum Mean Discrepancy (MMD) distances.
$\alpha $ is a hyperparameter that indicates the imbalance penalty.

Formally, we present the loss function of the proposed causal inference model as:
\begin{equation}
    \mathcal{L}^{\mbox{cau}} =   \mathcal{L}^{\mbox{fact}}  + \mathcal{L}^{\mbox{disc}}.
    \label{equ:cauloss}
\end{equation}

\subsection{Event Prediction with Causal Knowledge}
To improve the robustness of event predictions with imperfect real-world data, we incorporate causal information output by the causal inference model as priors to forecast future events. 
We introduce two robust learning modules into the training of event predictors:
(1) feature reweighting, which involves causal information to weight the original input features to obtain causally enhanced features, and (2) approximation constraints, which use the predicted potential outcomes as value range constraints applied to event prediction scores. 
Next, we introduce these two modules in detail.  
\subsubsection{\textbf{Feature Reweighting}}
Feature reweighting was introduced in object detection~\cite{kang2019few}, where a reweighting vector is learned to indicate the importance of meta features for detecting objects.
Here, we introduce a new feature reweighting method that leverages causal information. 
We use the ITE estimated from the causal inference model to reweight the event frequency features to predict future events.

\paragraph{Causal Feature Gates} 
We define a feature gate based on ITE calculated using predicted potential outcomes from the causal inference model. For the $j$-th treatment event, the estimated ITE of a location at time $t+\delta$ is as follows:
\begin{equation}
    \hat{\tau}_{(j)}^{t+\delta} = \hat{y}_{(j) }^{t+\delta}(1) - \hat{y}_{(j) }^{t+\delta}(0).
    \label{eq:ite-data}
\end{equation}
When considering multiple treatment events, we obtain the ITE vector $\hat{\bm{\tau}}^{t+\delta} \in \mathbb{R}^E$, where each element indicates a treatment event.
A linear layer $f_{\tau}$ with sigmoid function is then applied to model the association between the effects of different treatment events:
\begin{equation}
    \bm{\rho }^{t+\delta} = \sigma \big (f_{\tau}(\hat{\bm{\tau}}^{t+\delta}) \big ), 
\end{equation}
where $\bm{\rho}^{t+\delta} \in \mathbb{R}^E$ is the gating variables that will be applied to the original event frequency features. The sigmoid function $\sigma$ converts the gating variable into a soft gated signal with a range of $(0,1)$.

\paragraph{Reweighting Feature} 
We reweight the event frequency features using the gating variables defined above. 
It is worth emphasizing that the event frequency vector $\mathbf{x}^t$ has the same dimension as $\bm{\rho}^{t+\delta}$, and their corresponding elements represent the same event type. 
Nevertheless, we prefer not to apply the gating variables directly to the feature vector.
ITE examines whether the binary treatment variable affects the outcome of an instance, while the event frequency vector refers to discrete numbers.
To address this issue, we transform the event frequency feature into a latent vector using a position-wise feed-forward network (FFN)~\cite{vaswani2017attention}.
It maps the features into a continuous space, assuming that the gating variables can be aligned with the variables in this space. The formal procedures are defined as follows:
\begin{equation}
    \text{FFN} (\mathbf{x}^t ) = \text{ReLU} \big (\mathbf{x}^t\mathbf{W}^{r_0}+\mathbf{b}^{r_0} \big )\mathbf{W}^{r_1}+\mathbf{b}^{r_1},
\end{equation}
\begin{equation}
    \widetilde{\mathbf{x}}^t =\text{FFN} (\mathbf{x}^t ) \odot \bm{\rho}^{t+\delta} + \mathbf{x}^t, 
\end{equation}
where $\{\mathbf{W}, \mathbf{b}\}^{\{r_0,r_1\}}$ are learnable parameters.
A residual connection is added to ensure that the causally weighted elements still contain some original information. We denote the causality enhanced features across $\Delta$ historical steps as  $\widetilde{\mathbf{X}}^{\leq t} \in \mathbb{R}^{E \times \Delta}$.
Such features are fed into a predictor to perform event prediction, denoted as $\mathcal{P}(\widetilde{\mathbf{X}}^{\leq t},\mathbf{A})$.

\subsubsection{\textbf{Approximation Constraints}} The approximation constraints method was proposed to limit the reasonable range of the target variable during the model training process to generate a more robust model~\cite{muralidhar2018incorporating}. 
We follow this idea and propose a new method of integrating learned causal information into variable constraints.
Given an event predictor $\mathcal{P}$, we denote the model's event prediction for a location at time $t+\delta$ as $\hat{y}^{t+\delta}$.
Then, we assume that the causal range of the target variable, i.e., the event prediction, is $\hat{y}^{t+\delta} \in [l^{t+\delta},u^{t+\delta}]$.
The sample-wise boundaries are defined as:
\begin{equation}
    l^{t+\delta} = \mbox{Min} \big ( \hat{\mathbf{y}}^{t+\delta} \big ), \quad
    u^{t+\delta} = \mbox{Max}\big ( \hat{\mathbf{y}}^{t+\delta} \big ),
\end{equation}
where $\hat{\mathbf{y}}^{t+\delta} =\{ \hat{y}^{t+\delta}_{(j)}(1), \hat{y}^{t+\delta}_{(j)}(0) | 1 \leq j \leq  E \}  $ is the set of potential outcomes for all treatment events.
The minimum and maximum values are the lower and upper limits of the target variable for a given sample. Based on the range obtained from causal knowledge, we define a constraint loss term:
\begin{equation}
    \mathcal{L}^{\mbox{cstr}} = \sum_{t \in T}\sum_{i \in M} \text{ReLU} \big (l_{i}^{t+\delta}-\hat{y}_i^{t+\delta} \big ) + \text{ReLU} \big (\hat{y}_i^{t+\delta}-u_{i}^{t+\delta}  \big ) .
\end{equation}
The loss term can be involved during the training of the predictor $\mathcal{P}$. Given the proposed robust learning modules for event prediction, we train the predictor by minimizing the following loss function:
\begin{equation}
    \mathcal{L}^{\mbox{evt}} = \mathcal{L}^{\mbox{pred}} + \mu \cdot \mathcal{L}^{\mbox{cstr}},
\label{equ:eventloss}
\end{equation}
where $ \mathcal{L}^{\mbox{pred}} $ is the loss function defined by the predictor $\mathcal{P}$ and $\mu$ is a hyperparameter. 
The training steps of the proposed method are shown in Algorithm~\ref{alg:event}.
\begin{algorithm}
\DontPrintSemicolon
\KwIn{Observational event data $\mathcal{D} = \big \{  \{\mathbf{X}^{\leq t}_i,\mathbf{c}^{\leq t}_i, y^{t+\delta}_i, \}^{M}_{i=1}, \mathbf{A} \big \}_{t \in T}$, a predictor $\mathcal{P}$ with randomly initialized parameters, and initialized model \ours, including causal inference model  $\mathcal{C}$ and robust learning modules $\mathcal{R}$ for event prediction.}
\Comment*[l]{Train the causal inference model $ \mathcal{C} $.}
\While{$\mathcal{C} $ has not converged }{
Input $\mathcal{D}$ to $\mathcal{C}$ to obtain predicted potential outcomes. Calculate the loss $\mathcal{L}^{\mbox{cau}}$.\;
Update  $\mathcal{C}$ by optimizing Eq.~\ref{equ:cauloss}. 
}
Freeze the causal inference model $\mathcal{C}$.\;
\Comment*[l]{Train the predictor $\mathcal{P}$ and robust modules $\mathcal{R}$ with $\mathcal{C}$.}
\While{$\mathcal{P},\mathcal{R}$ have not converged}{
Input $ \mathcal{D}$ to  $\mathcal{C}$ to obtain predicted potential outcomes.\;
Calculate causally reweighed features $\widetilde{\mathbf{X}}$ from robust learning modules $\mathcal{R}$ using predicted potential outcomes.\;
Input $\{ \widetilde{\mathbf{X}}, \mathbf{A}\}$ to the predictor $\mathcal{P}$ to obtain the event prediction. Calculate the prediction loss $\mathcal{L}^{\mbox{pred}}$. \;
Calculate the constraint loss $\mathcal{L}^{\mbox{cstr}}$ from $\mathcal{R}$ using predicted potential outcomes.\;
Update the predictor $\mathcal{P}$ and causal modules $\mathcal{R}$ by optimizing Eq.~\ref{equ:eventloss}. 
}
\KwOut{the updated model \ours.}
\caption{\textbf{CAPE}}
\label{alg:event}
\end{algorithm}

\section{Experimental Evaluation}
\label{sec:question}
The goal of the experimental evaluation is to answer the following research questions:
\textbf{RQ1}: How well does \ours~ estimate ITEs in observational event data? 
\textbf{RQ2}: Can \ours~ improve the robustness of event prediction models? \textbf{RQ3}: What causal information can we learn from studies of causally related event prediction?

Next, we will describe the experimental setup and then show the experimental results to address the above questions.

\subsection{Datasets}
Experimental evaluation is conducted on two data sources: Integrated Conflict Early Warning System (ICEWS)~\cite{icews}, and Global Database of Events, Language, and Tone (GDELT)~\cite{leetaru2013gdelt}.
These two data sources include daily events encoded from news reports.\footnote{For event data from GDELT, we only select root events identified in news reports.}
We construct event datasets for four countries, i.e., \textbf{India}, \textbf{Nigeria}, \textbf{Australia} and \textbf{Canada}, based on their large volume of events. Events are categorized into 20 main categories (e.g., appeal, demand, protest, etc.) according to CAMEO methodology~\cite{DVN/28075/SCJPXX_2015}. Each event is encoded with geolocation, time (day, month, year), category, etc. 
In this work, we focus on predicting one category of events: \textit{protest}, as the target variable, and using event historical data of all event types as feature variables. 
Data statistics are shown in Table~\ref{sum-data}. Positive in the table refers to the proportion of positive samples.
\begin{table}[]
\caption{Dataset Statistics. 
 $M$ is the number of locations in each dataset. \textbf{Positive} indicates the ratio of positive samples, i.e., the protest event has occurred.
For India, we select top locations based on the total number of events. \textbf{Location} represents the geographical level of events. 
}
\label{sum-data}
\begin{center}
\begin{tabular}{lrr cccc}
\toprule
\textbf{Dataset} &  \textbf{$M$} & \textbf{Positive} & \textbf{Location} & \textbf{Time} & \textbf{Time Unit}&  \textbf{Source} \\
\midrule
\textbf{India}  &  14 & 30.1\% & State & 2000-2017 & 3 days & ICEWS\\
\textbf{Nigeria}  &  6  & 65.7\% & Geopolitical zone & 2015-2020 & 1 day& GDELT\\
\textbf{Australia} & 8  & 44.4\% & State  &2015-2020 & 1 day&GDELT\\
\textbf{Canada} & 13   & 26.8\% & State &2015-2020 & 1 day&GDELT\\
\bottomrule
\end{tabular}
\end{center}
\end{table} 

\subsection{Evaluation Metrics}

For the ITE estimation, since there is no ground truth counterfactual outcomes,
we report the ATT error $\epsilon_{\text{ATT}} = |\text{ATT}-\mathbb{E}_{\mathcal{S}}[\hat{y}^{t+\delta}(1)-\hat{y}^{t+\delta}(0)|c^{\leq t}=1]|$~\cite{shalit2017estimating}, where ATT is the true average treatment effect on the treated, i.e., $\text{ATT} = \mathbb{E}_{\mathcal{S}}[y^{t+\delta}|c^{\leq t}=1]-\mathbb{E}_{\mathcal{S}}[y^{t+\delta}|c^{\leq t}=0]$.
$\mathcal{S}$ denotes the subset of samples simulating a randomized controlled trial.
Specifically, given the treatment event, we employ a 1-nearest neighbor algorithm~\cite{yang2006distance} to find a matching control instance (without replacement) for each treated instance. Euclidean distance is adopted to measure feature vectors.
The matching process is performed for each location.

We quantify the predictive performance of event prediction based on Balanced Accuracy (BACC), i.e, $\text{BACC} = (\text{TPR}+\text{TNR})/2$. TPR and TNR are the true positive rate and true negative rate, respectively. BACC is a good metric when the classes are imbalanced.

\subsection{Comparative Methods}
For the ITE estimation, we compare our causal inference model, notated as $\ours_{\mathcal{C}}$, with two groups of baselines: (i) Tree based methods: Bayesian Additive Regression Trees (\bart)~\cite{chipman2010bart} and Causal Forest (\cf)~\cite{wager2018estimation};
    (ii) Representation learning based methods: Counterfactual regression with MMD (\cfrmmd)~\cite{shalit2017estimating} and Wasserstein metric (\cfrwass)~\cite{shalit2017estimating}, 
    Causal Effect Variational Autoencoder (\cevae)~\cite{louizos2017causal}, Network Deconfounder  (\netdeconf)~\cite{guo2019learning}, and Similarity Preserved Individual Treatment Effect (\site)~\cite{yao2018representation}.
 
We study three variants of our model to examine the impact of different components in our model: (i) $\ours_{\mathcal{C}-G}$ which removes the spatial feature learning. (ii) $\ours_{\mathcal{C}-T}$ which replaces the temporal feature learning with a simple linear transformation. (iii) $\ours_{\mathcal{C}-B}$ removes the loss term $\mathcal{L}^{\mbox{disc}}$.

To evaluate the effectiveness of proposed robust learning modules in event prediction, we adopt two spatiotemporal models as the predictor $\mathcal{P}$: 
(i) \cola~\cite{deng2020cola}: A graph-based framework for long-term Influenza-like illness prediction; 
(ii) \gwnet~\cite{wu2019graph}: A state-of-the-art spatiotemporal graph model for traffic prediction.
Given the spatiotemporal characteristics of societal event data, these models can be well applied to our problem. 
Note that we do not adopt protest event prediction models~\cite{deng2019learning,deng2021understanding} because they model on more complex data, such as text and knowledge graphs. 
We leave the causal exploration of such complex data to future work.

\section{Implementation Details}
For the causal inference model, we use three gated temporal convolutional layers with dilation factors $d=1,2,4$, and two graph convolutional layers. The dimension $d_a$ is set to 10. 
The feature dimensions of all other hidden layers including $d_s$ are set to be equal and searched from $\{16,32,64\}$.

The number of treatment events $E$ is 20, where each treatment event corresponds to an event type, such as appeal and protest.
Following previous work~\cite{deng2019learning, 10.1145/3394486.3403209}, we set the historical window size $\Delta$ to 7 and the lead time $\delta$ to 1.
The hyperparameter $\eta$ used for parameter regularization is fixed to 1e-5.
We use the squared linear MMD for representation balancing~\cite{shalit2017estimating}.
The imbalance penalty $\alpha$ is searched from $\{10^{-5},10^{-4},10^{-3},10^{-2},10^{-1}\}$. 
The scaling term $\mu$ in Eq.~\ref{equ:eventloss}
is searched from $\{10^{-5},10^{-4},10^{-3},10^{-2},10^{-1}\}$. All parameters are initialized with Glorot initialization \cite{glorot2010understanding} and trained using the Adam \cite{kinga2015method} optimizer with learning rate $10^{-3}$ and dropout rate 0.5. The batch size is set to 64. 
We use the objective value on the validation set for early stopping.

For causal inference baselines,
\cf~\footnote{\url{https://rdrr.io/cran/grf/man/causal_forest.html}} and \bart~\footnote{\url{https://rdrr.io/cran/BART/}} are implemented using R packages.
We implement the causal inference models \cfrmmd, \cfrwass, \site~ by ourselves and use the source code of \cevae~\footnote{\url{https://github.com/rik-helwegen/CEVAE_pytorch}} and \netdeconf~\footnote{\url{https://github.com/rguo12/network-deconfounder-wsdm20}}. We apply parameter searching on all baseline models. For representation learning based approach, the dimension of hidden layers are searched from $\{32,64,128\}$ and the number of hidden layers are searched from $\{1,2\}$. For models that introduce balancing representation learning, we search the hyperparameter from $\{10^{-5},10^{-4},10^{-3},10^{-2},10^{-1}\}$. The model \netdeconf~ involves an auxiliary network and we use the geographic adjacency matrix for locations.

For the experiments on event forecasting, we run the source code of \cola~\footnote{\url{https://github.com/amy-deng/colagnn}} and \gwnet~\footnote{\url{https://github.com/nnzhan/Graph-WaveNet}}. For event prediction models, we fixed the dimension of hidden layers to 32.
\cola~ takes the geographic adjacency matrix as input and \gwnet~ learns the adaptive adjacency matrix.

We report the average of 5 randomized trials for all experiments.
At each training, we randomly split the data into training, validation, and test sets at a ratio of 70\%-15\%-15\% with a fixed seed value. All python code is implemented using Python 3.7.7 and Pytorch 1.5.0 with CUDA 9.2.

\section{Experimental Results}

\subsection{Results of ITE Estimation (RQ1)}

\begin{table}[]
\centering
\caption{ITE estimation results showing the mean and standard deviation of ATT errors on all datasets with treatment event being \textit{Appeal}.
Lower is better. 
}
\label{tab:causal-result}
\scalebox{1.0}{\begin{tabular}{lcccc}
\toprule
        & \multicolumn{4}{c}{Treatment event: Appeal}                   \\
    \midrule
        & India         & Nigeria       & Australia     & Canada        \\
\midrule
\bart    & $.114 \pm .012$   & $.024 \pm .015$   & $.034 \pm .018$   & $.035 \pm .017$   \\
\cf      & $.112 \pm .011$   & $.034 \pm .015$   & $.038 \pm .021$   & $.038 \pm .016$   \\
\cfrmmd  & $.014 \pm .012$   & $.020 \pm .013$   & $.016 \pm .011$   & $.017 \pm .010$    \\
\cfrwass & $.021 \pm .010$   & $.025 \pm .016$   & $.019 \pm .013$   & $.012 \pm .009$   \\
\cevae   & $.012 \pm .007$   & $.018 \pm .014$   & $.018 \pm .011$   & $.017 \pm .016$   \\
\site    & $.013 \pm .009$   & $.019 \pm .015$   & $.012 \pm .008$   & $.016 \pm .013$   \\
\netdeconf  & $.034 \pm .027$   & $.030 \pm .017$   & $.022 \pm .023$   & $.034 \pm .020$    \\
\midrule
$\ours_{\mathcal{C}}$     & $\mathbf{.011 \pm .008}$ & $\mathbf{.010 \pm .007}$ & $\mathbf{.010 \pm .007}$ & $\mathbf{.008 \pm .006}$ \\
$\ours_{\mathcal{C}-G}$   & $.019 \pm .009$ & $.019 \pm .011$ & $.019 \pm .008$ & $.016 \pm .011$ \\
$\ours_{\mathcal{C}-T}$   & $\mathbf{.011 \pm .009}$ & $.027 \pm .021$ & $.013 \pm .010$ & $.011 \pm .007$ \\
$\ours_{\mathcal{C}-B}$ & $.018 \pm .012$ & $.026 \pm .013$ & $.020 \pm .013$ & $.019 \pm .012$ \\
\bottomrule
\end{tabular}}
\end{table}

\begin{table}[]
\centering
\caption{ITE estimation results showing the mean and standard deviation of ATT errors on all datasets with treatment event being \textit{Reject}.
Lower is better. 
}
\label{tab:causal-result-2}
\begin{tabular}{lcccc}
\toprule
       & \multicolumn{4}{c}{Treatment event: Reject}       \\
        \midrule
       & India       & Nigeria     & Australia   & Canada      \\
        \midrule
\bart    & $.184 \pm .012$ & $.072 \pm .021$ & $.045 \pm .030$ & $.070 \pm .021$ \\
\cf      & $.181 \pm .013$ & $.062 \pm .021$ & $.039 \pm .031$ & $.067 \pm .022$ \\
\cfrmmd  & $.022 \pm .011$ & $.021 \pm .020$ & $.024 \pm .013$ & $.018 \pm .009$ \\
\cfrwass & $.016 \pm .012$ & $.018 \pm .009$ & $.026 \pm .021$ & $.016 \pm .011$ \\
\cevae   & $.015 \pm .010$ & $.021 \pm .018$ & $.020 \pm .012$ & $.019 \pm .015$ \\
\site    & $\mathbf{.010 \pm .008}$ & $.019 \pm .010$ & $.022 \pm .019$ & $.024 \pm .012$ \\
\netdeconf  & $.026 \pm .021$ & $.020 \pm .018$ & $.024 \pm .019$ & $.023 \pm .014$ \\
\midrule
$\ours_{\mathcal{C}}$   & $.016 \pm .011$ & $.016 \pm .005$ & $.012 \pm .009$ & $.015 \pm .012$ \\
$\ours_{\mathcal{C}-G}$ & $\mathbf{.010 \pm .013}$ & $.020 \pm .012$ & $.015 \pm .010$ & $\mathbf{.011 \pm .011}$ \\
$\ours_{\mathcal{C}-T}$ & $.015 \pm .014$ & $\mathbf{.014 \pm .013}$ & $\mathbf{.007 \pm .004}$ & $.016 \pm .011$ \\
$\ours_{\mathcal{C}-B}$ & $.017 \pm .013$ & $.028 \pm .013$ & $.018 \pm .014$ & $.018 \pm .009$ \\
\bottomrule
\end{tabular}
\end{table}

\label{sec:ite}
To evaluate the effectiveness of our proposed causal inference framework, we limit the number of treatment events to be one and compare our model with other baselines. 
We focus on two treatment events: \textit{appeal} and \textit{reject}.
The motivation is that \textit{appeal} events might be a potential cause of protest events, as they express a serious or urgent request, typically to the public. \textit{Reject} events represent verbal conflicts~\cite{DVN/28075/SCJPXX_2015}, which contain dissatisfaction with the current state and may lead to a future occurrence of protest.
Table~\ref{tab:causal-result} and Table~\ref{tab:causal-result-2} report the ATT errors of all causal inference models on four datasets when treatment variable to be \textit{appeal} and \textit{reject}, respectively. 
The results show that the tree-based model performs worse than the representational learning-based model.
The findings reflect the limitations of the tree-based models and highlight the benefits of representation learning for estimating ITE for observational event data.
\cfrmmd~ and \cfrwass~ learn a balanced representation such that the induced treated and control distributions look similar. 
Both models achieved good results in most cases, demonstrating the importance of controlling for representation distributions to predict potential outcomes.
\cevae~  learns latent variables based on variational autoencoders and \site~ focuses on capturing local similarities to estimate ITE.
These two models present the most stable and relatively small ATT errors in all settings. This suggests that learning the latent variables and considering similarity information is useful for estimating ITE for observational event data.
The model \netdeconf~ learns hidden confounders by leveraging network/spatial information. However, it does not outperform representation-based baselines.
This may be because the model was designed for semi-synthetic datasets and the spatial characteristics of observational event data are different from the network used in the original paper.
Our proposed causal inference framework learns hidden confounders while capturing spatial and temporal information and achieves the best performance.
For our model variants, we observe that removing the representation balancing makes the results worse. Ignoring the temporal or spatial feature learning can also deteriorate the results.
This reflects the possible spatiotemporal dependencies underlying the hidden confounders. It also demonstrates the capability of the proposed model in capturing the spatiotemporal information of the observational event data.

\subsection{Robustness Tests in Event Prediction (RQ2)}
In this subsection, we perform two robustness tests on event prediction for all datasets and conduct a case study on the proposed feature reweighting module.

\subsubsection{\textbf{Robustness to Test Noise}} 
\begin{figure}[t]
  \centering 
\begin{subfigure}{.25\textwidth}
    \centering
    \includegraphics[width=1\linewidth]{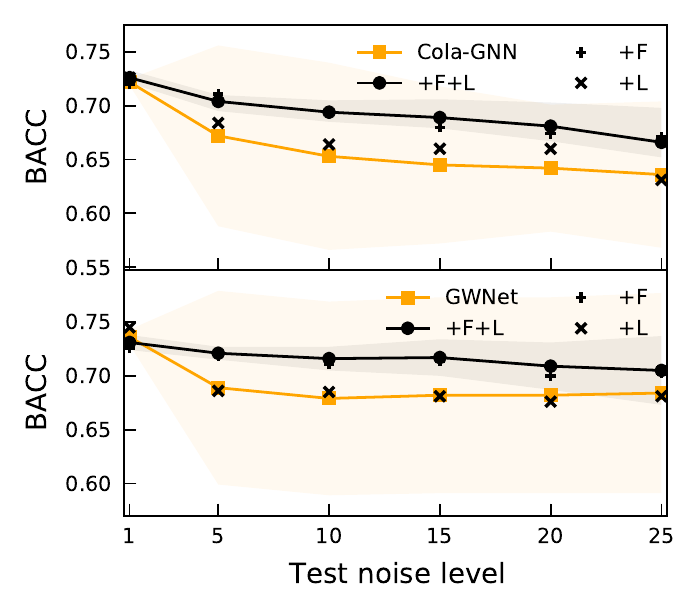}
    \caption{\textbf{India}}\label{fig:test-noise-ind}
  \end{subfigure} 
 \hspace{-5pt}
  \begin{subfigure}{.25\textwidth}
    \centering
    \includegraphics[width=1.0\linewidth]{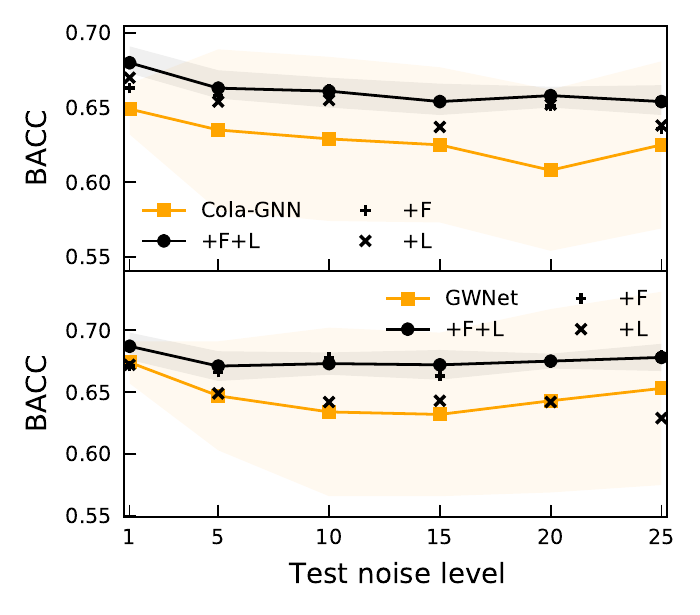}
    \caption{\textbf{Nigeria}}
  \end{subfigure} 
  \hspace{-5pt}
  \begin{subfigure}{.25\textwidth}
    \centering
    \includegraphics[width=1.0\linewidth]{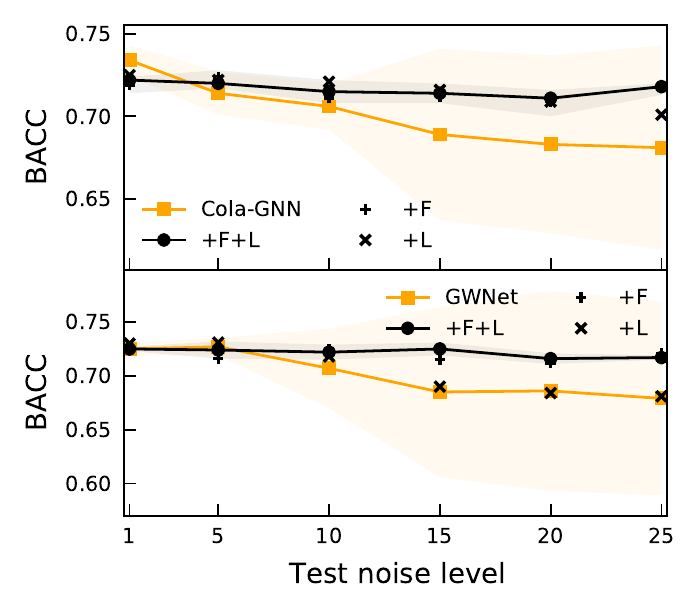}
    \caption{\textbf{Australia}}
  \end{subfigure}  
  \hspace{-5pt}
  \begin{subfigure}{.25\textwidth}
    \centering
    \includegraphics[width=1.0\linewidth]{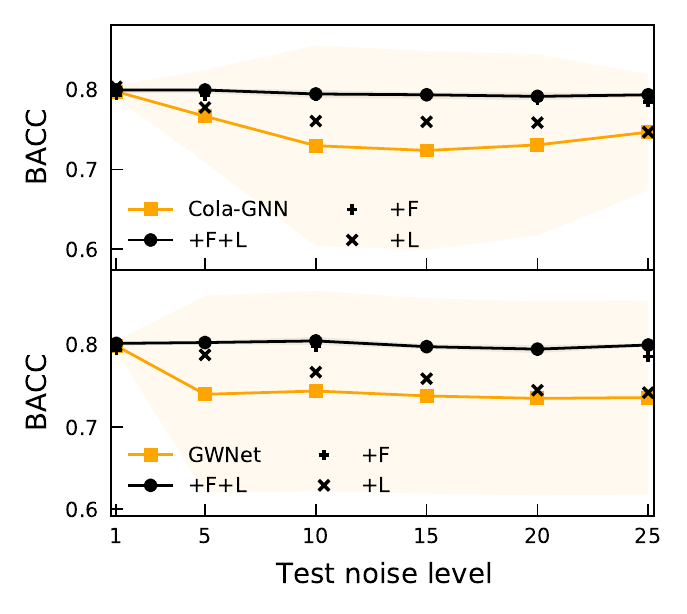}
    \caption{\textbf{Canada}}
  \end{subfigure} 
 \caption{BACC value of event prediction when varying the noise level in validation and test sets. Higher is better.} 
  \label{fig:robust-test-noise}
 \end{figure}

A model is considered to be robust if its output variable is consistently accurate when one or more input variables drastically change due to unforeseen circumstances. 
In this setting, we add Poisson noise into the validation and test sets while keeping the training data noise-free. 
We aim to verify whether our method guarantees good prediction performance when the test input features are biased.
We vary the rate parameter (aka expectation) of the Poisson distribution from 1 to 25 and provide the comparison results for different noise levels in Fig~\ref{fig:robust-test-noise}.
We notice that training with the proposed robust learning module leads to higher average BACC results and lower variance over multiple runs.
In most cases, the feature reweighting module (+F) contributes more in improving the prediction performance.
Incorporating these two modules (+F+L) can lead to better overall results.
The results suggest that forecasting events with learned causal information is beneficial to improve the robustness of the prediction.

\subsubsection{\textbf{Robustness to Training Noise}}
\begin{figure}[t]
  \centering 
\begin{subfigure}{.25\textwidth}
    \centering
    \includegraphics[width=1\linewidth]{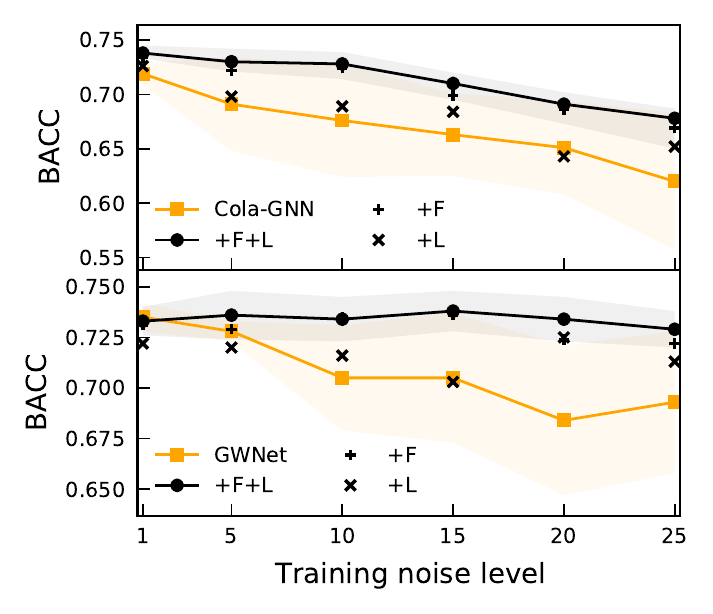}
    \caption{\textbf{India}}
  \end{subfigure} 
 \hspace{-5pt}
  \begin{subfigure}{.25\textwidth}
    \centering
    \includegraphics[width=1\linewidth]{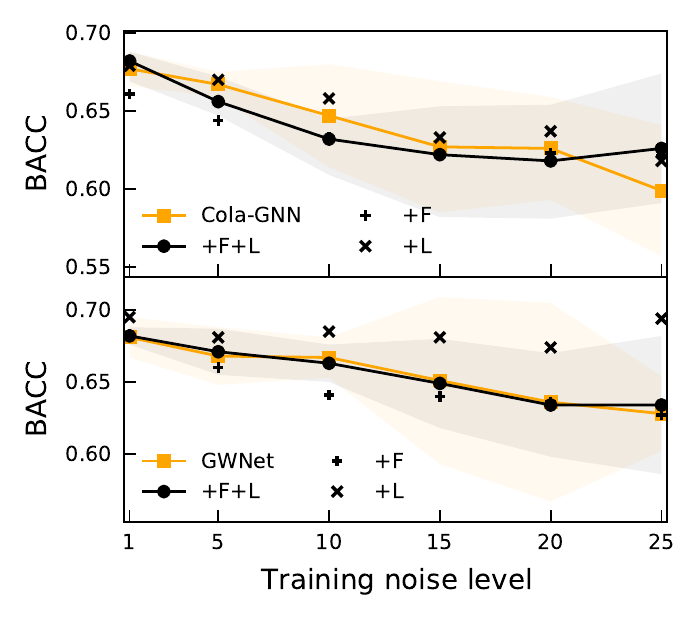}
    \caption{\textbf{Nigeria}}\label{fig:test-ni}
  \end{subfigure} 
  \hspace{-5pt}
  \begin{subfigure}{.25\textwidth}
    \centering
    \includegraphics[width=1.0\linewidth]{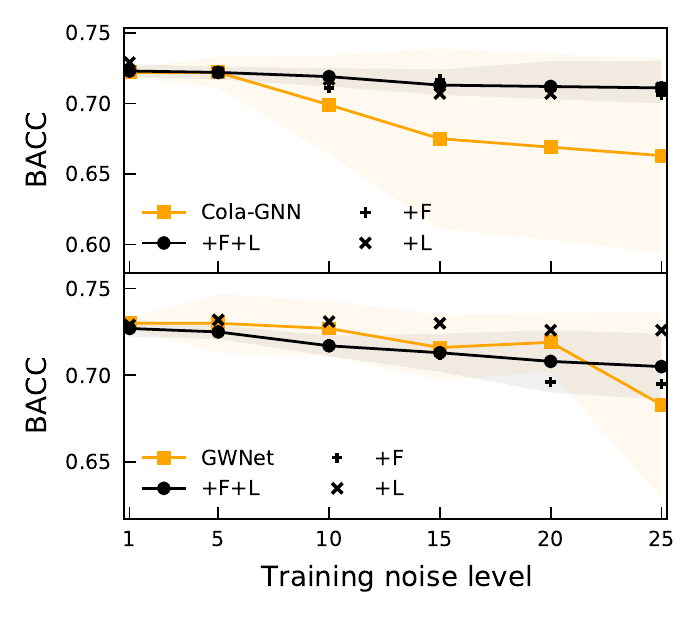}
    \caption{\textbf{Australia}}\label{fig:test-as}
  \end{subfigure}  
  \hspace{-5pt}
  \begin{subfigure}{.25\textwidth}
    \centering
    \includegraphics[width=1\linewidth]{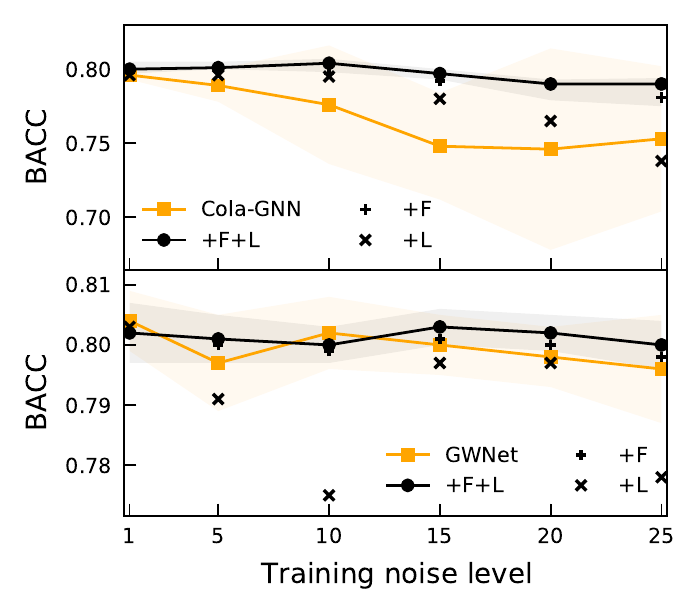}
    \caption{\textbf{Canada}}
  \end{subfigure} 
 \caption{BACC value of event prediction when varying the noise level in the training set. Higher is better.
 } 
  \label{fig:robust-train-noise}
 \end{figure}

Human errors or machine failures in real-world data collection usually reduce data accuracy.
With this motivation, we assume that only the training data are biased and test whether our method can achieve decent event prediction results on unbiased test data.
As shown in Fig.~\ref{fig:robust-train-noise}, applying robust learning modules can help the prediction model achieve better performance in BACC when the noise level increases. 
Adding the approximation constraint loss (+L) can lead to a higher BACC than adding the two modules (Fig.~\ref{fig:test-ni} and Fig.~\ref{fig:test-as}). 
The results also illustrate that even with biased data (with corrupted features), the trained causal inference model learns valuable information that contributes to event prediction.

\subsubsection{Case Study of Feature Reweighting}
\begin{figure}
  \centering 
\begin{subfigure}{.25\textwidth}
    \centering
    \includegraphics[width=1\linewidth]{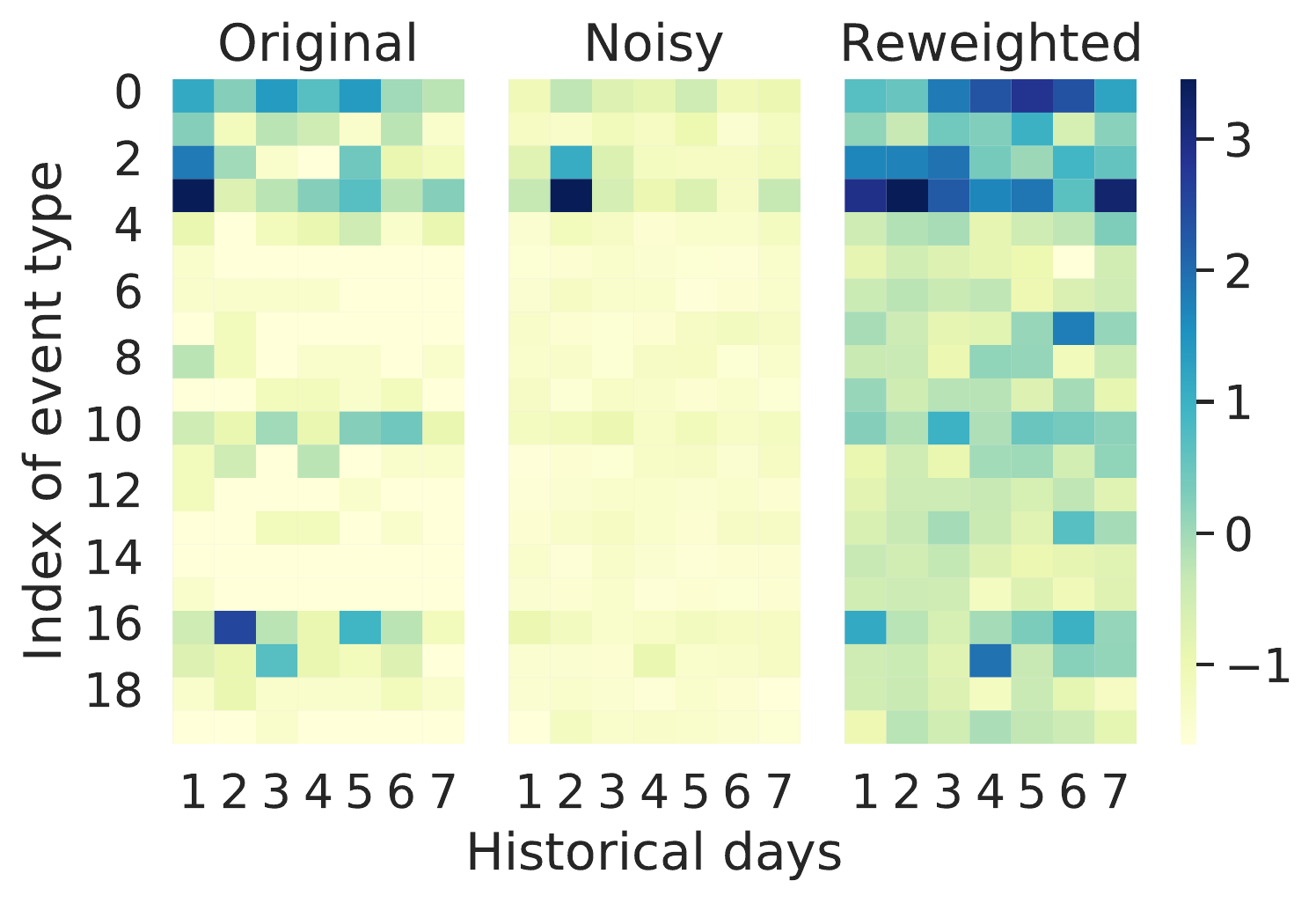}
    \caption{Sample 1}
  \end{subfigure} 
 \hspace{-5pt}
  \begin{subfigure}{.25\textwidth}
    \centering
    \includegraphics[width=1\linewidth]{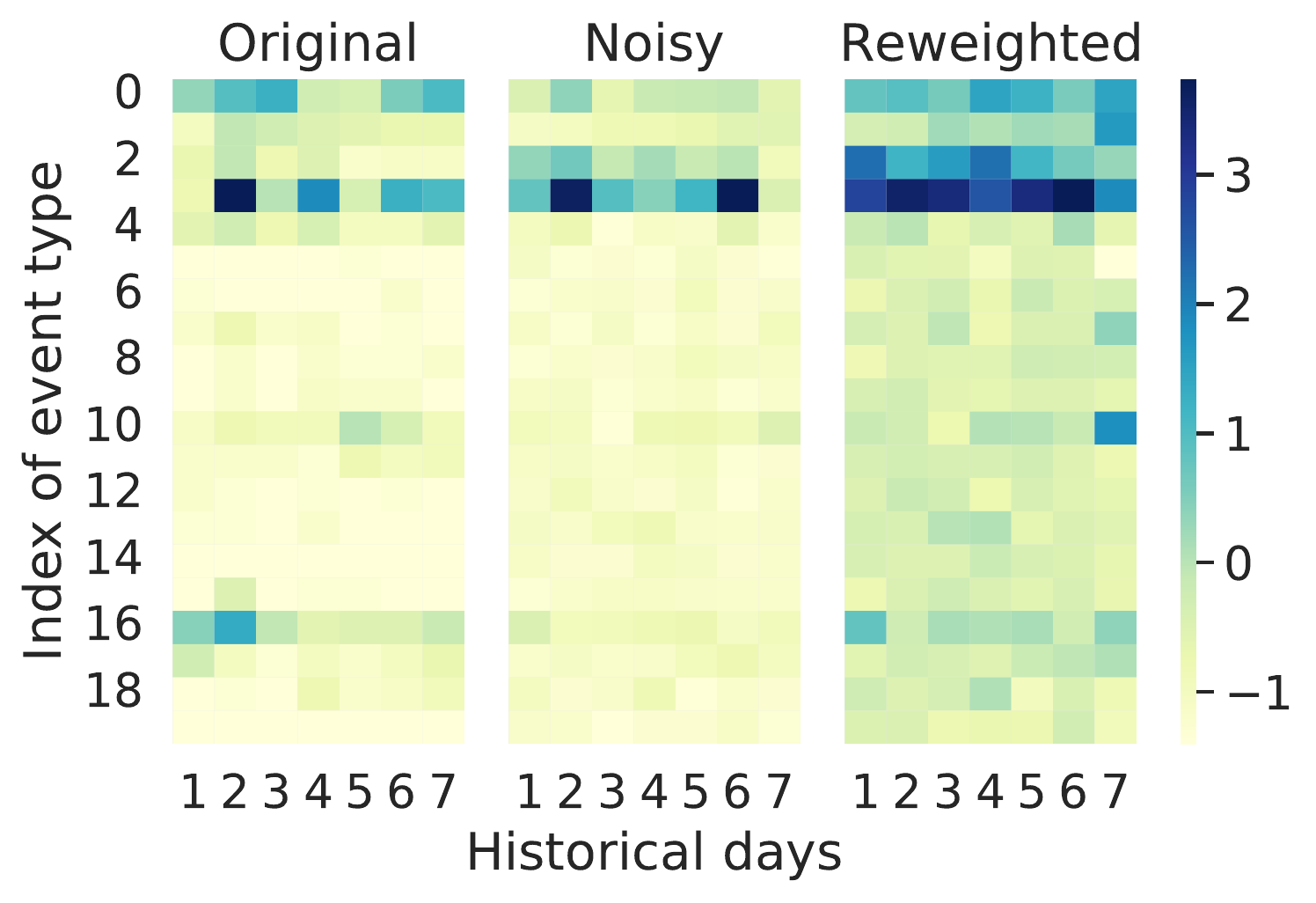}
     \caption{Sample 2}
  \end{subfigure} 
   \hspace{-5pt}
  \begin{subfigure}{.25\textwidth}
    \centering
    \includegraphics[width=1.0\linewidth]{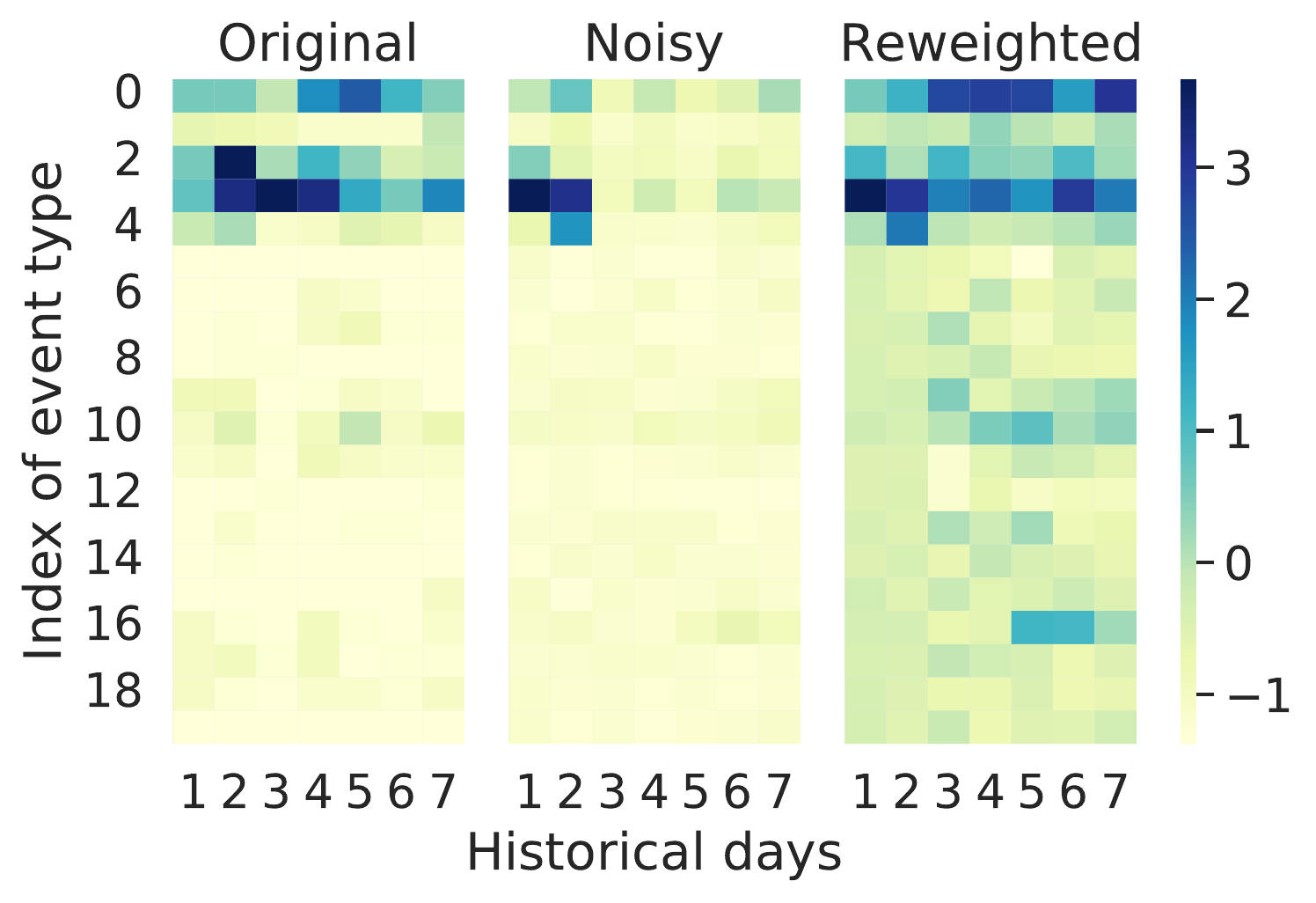}
    \caption{Sample 3}
  \end{subfigure}  
  \hspace{-5pt}
   \begin{subfigure}{.25\textwidth}
    \centering
    \includegraphics[width=1\linewidth]{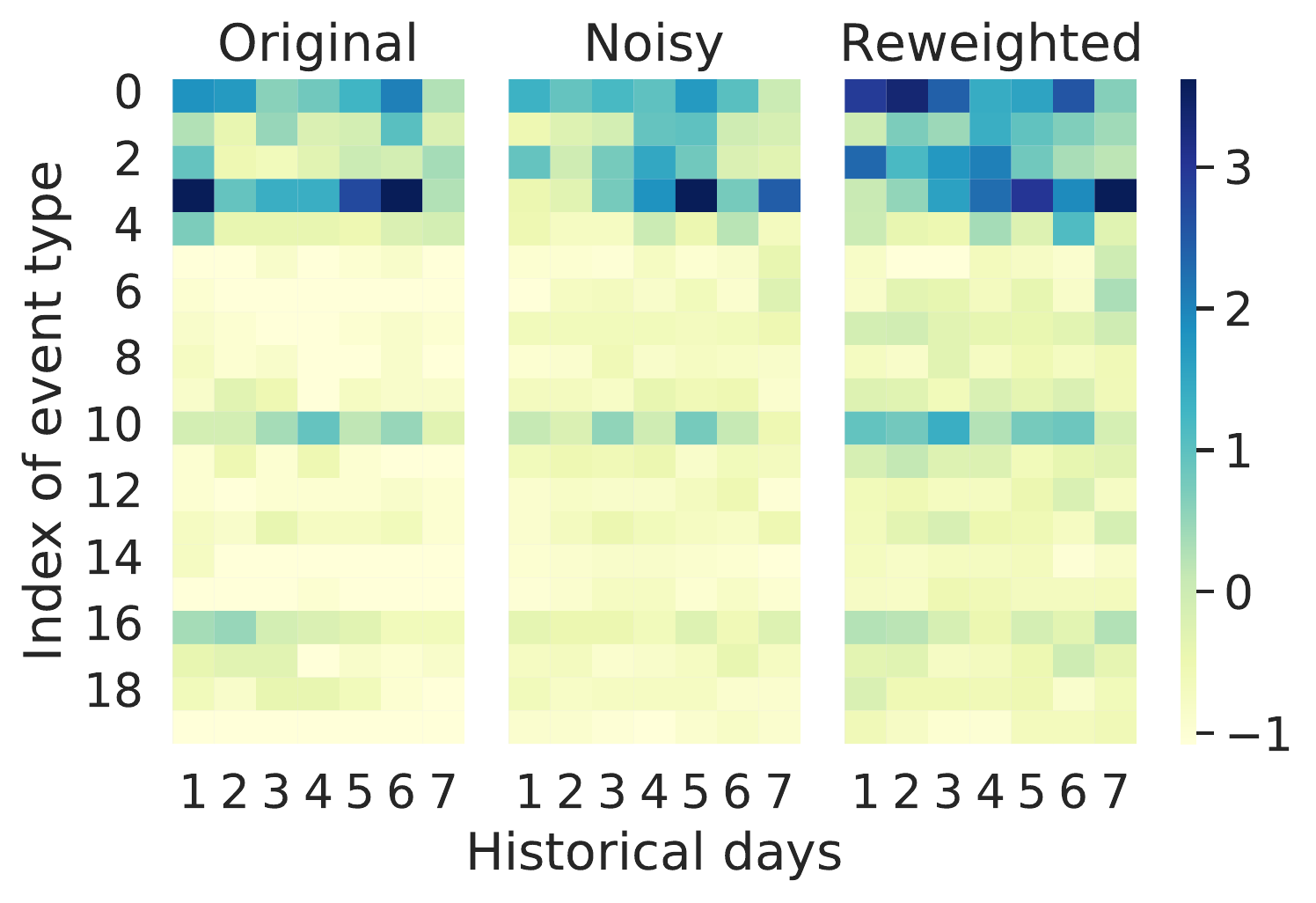}
     \caption{Sample 4}
  \end{subfigure} 
 \caption{Distributions of original, noisy, and reweighted features of corrupted samples on the India dataset.
 } 
 \label{fig:feature-map}
 \end{figure}

To illustrate the functionality of the proposed feature reweighting on robust event prediction, we provide several examples in the India dataset, as shown in Fig.~\ref{fig:feature-map}.
We use \cola~ for analysis, given the more apparent improvements when it is applied with the feature reweighting module.
Specifically, we first train an event prediction model on the India dataset using \cola~ with the feature reweighting module. 
We select four corrupted test samples with random noise added to their input features (noise level of 5). 
We visualize the original features, the noisy features, and the ones obtained from the feature reweighting module.
We can observe that the reweighted features can encode similar patterns of original features. 
It highlights the advantages of the ITE used in the feature reweighting module and demonstrates its ability to capture crucial information underlying the data distribution.

\subsection{Causal Effect in Societal Events (RQ3)}
\begin{figure}[t]
  \centering 
\begin{subfigure}{.25\textwidth}
    \centering
    \includegraphics[width=1\linewidth]{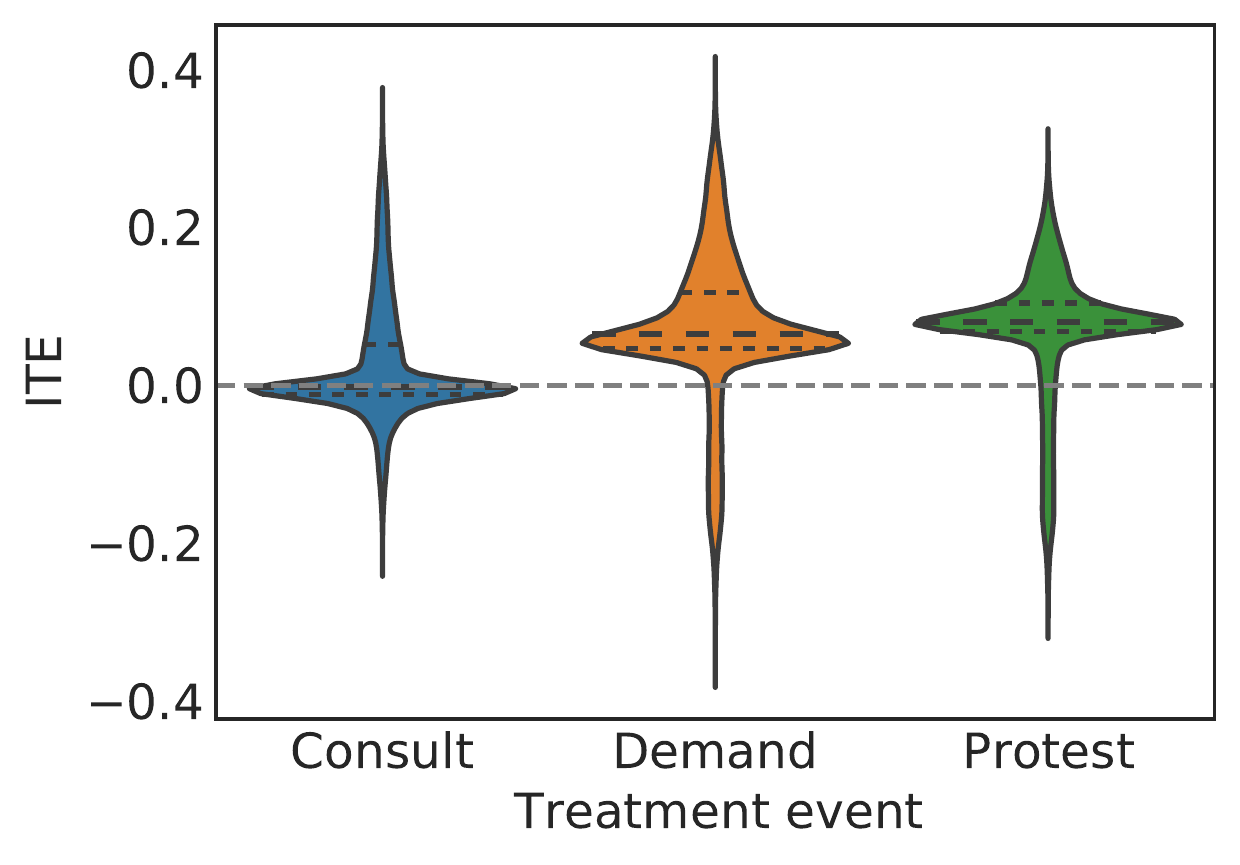}
    \caption{\textbf{India}}
  \end{subfigure} 
 \hspace{-5pt}
  \begin{subfigure}{.25\textwidth}
    \centering
    \includegraphics[width=1\linewidth]{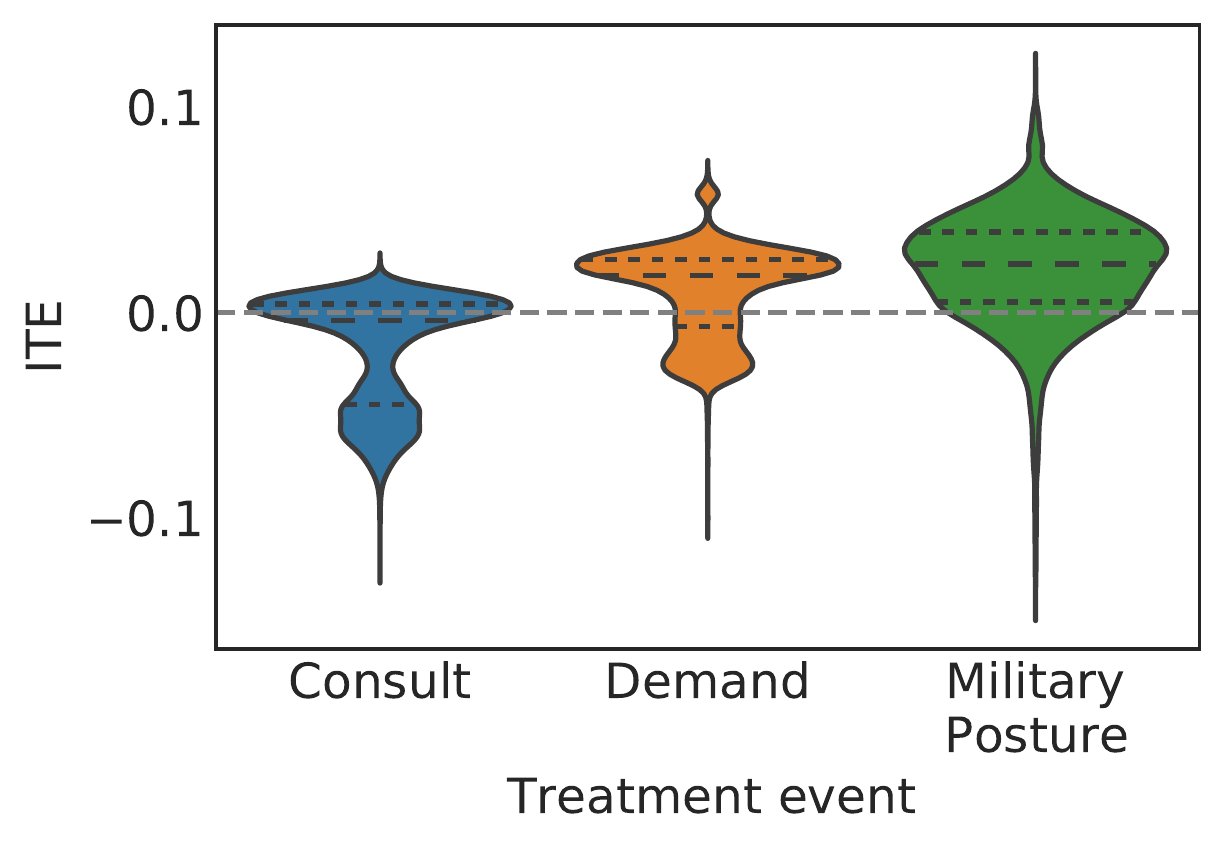}
    \caption{\textbf{Nigeria}}
  \end{subfigure} 
  \hspace{-5pt}
  \begin{subfigure}{.25\textwidth}
    \centering
    \includegraphics[width=1.0\linewidth]{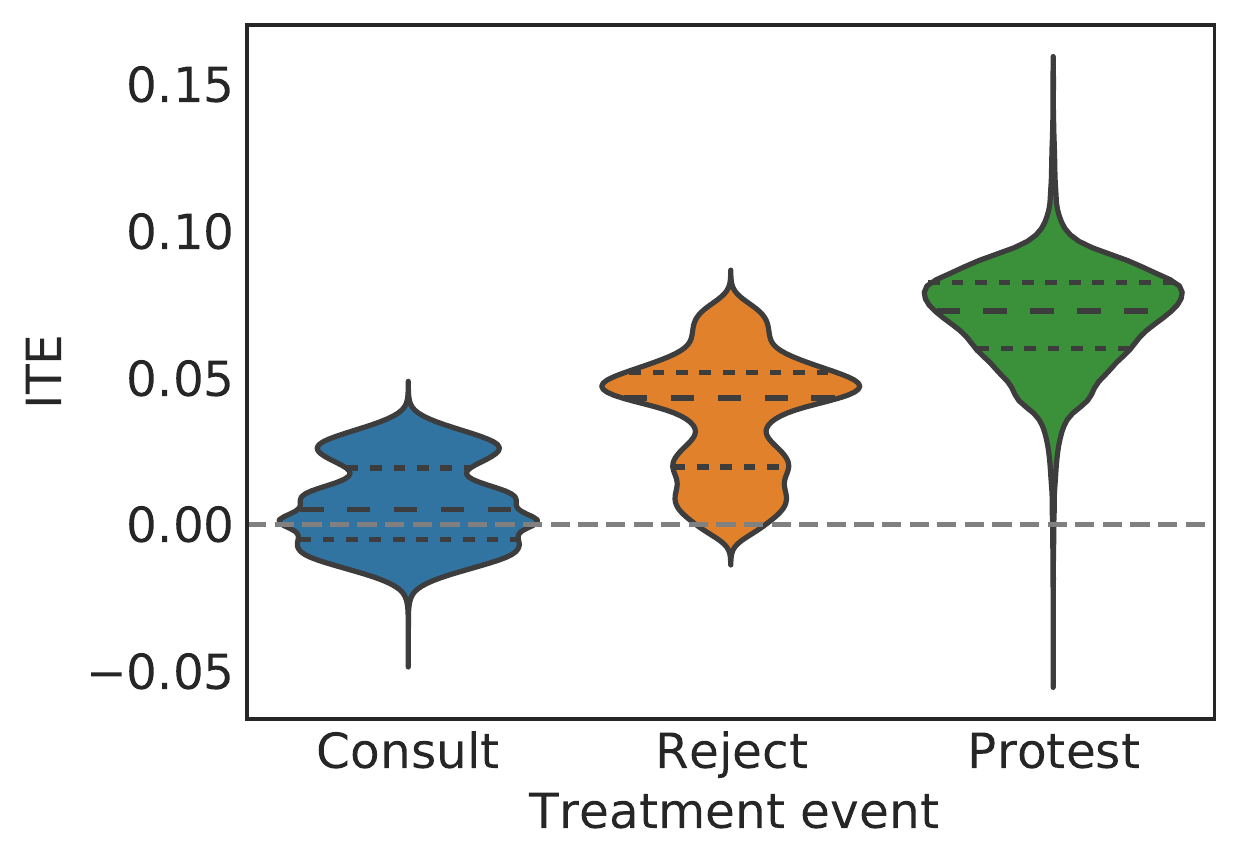}
    \caption{\textbf{Australia}}
  \end{subfigure}  
  \hspace{-5pt}
  \begin{subfigure}{.25\textwidth}
    \centering
    \includegraphics[width=1\linewidth]{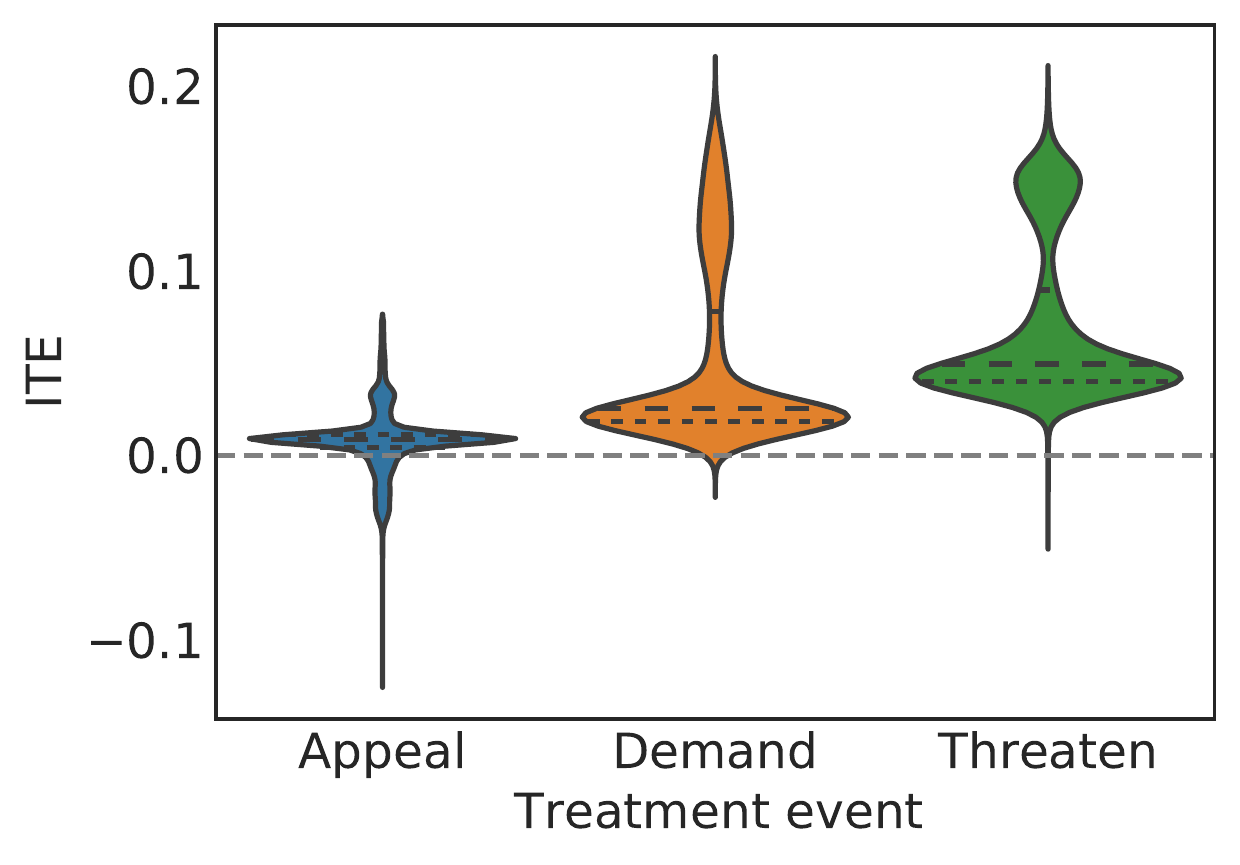}
    \caption{\textbf{Canada}}
  \end{subfigure} 
 \caption{ITE distribution of different treatment events on the outcome \textit{protest} in violin plot. 
 The loosely dashed line represents the median.
 } 
  \label{fig:ite-box}
 \end{figure}

In our study, whether there is a significant increase in certain types of events (e.g., appeal) over the past window is defined as the treatment of a location.
The outcome is the future occurrence of a target event, i.e., protest.
In this case, the ITE measures the difference in the outcome of the protest occurring between the two scenarios of the treatment event (i.e., increased or not).
Thus, when the necessary assumptions hold, it implies a causal effect of the treatment event on the protest.
A higher ITE suggests that an increase in a treatment event will be more influential in the occurrence of future protests, compared to a decrease or no change.
To better illustrate the effect of treatment events on future protests, we visualize the predicted ITEs based on Eq.~\ref{eq:ite-data}.
Violin plots for the four datasets are shown in Fig.~\ref{fig:ite-box}. 
We select three treatment events for each dataset. They have relatively low, moderate, and high ITE on average, respectively.
The results vary from datasets due to different social environments. 
In India and Australia, massive historical protests may lead to future protests. 
In Nigeria and Canada, events related to military posture and threats, respectively, are likely to be more dominant factors in future protests.
Nevertheless, we hardly conclude that protests will occur when the treatment event increases substantially because both types of events can be affected by hidden variables (i.e., unknown social factors).
These results can provide supporting evidence for conjectures on protest triggers and generate hypotheses for future experiments.

\section{Conclusion and Future Work}
Learning causal effects of societal events is beneficial to decision-making and helps practitioners understand the underlying dynamics of events. 
In this paper, we introduce a deep learning framework that can estimate the causal effects of societal events and predict societal events simultaneously.
We design a novel spatiotemporal causal inference model for estimating ITEs and propose two robust learning modules that use the learned causal information as prior knowledge for societal event prediction.
We conducted extensive experiments on several real-world event datasets and showed that our approach achieves the best results in ITE estimation and robust event prediction.
One future direction is to examine other potential causes of event occurrence, such as events with specific themes and potentially biased media coverage. 

\section{Broader Impacts}
This work aims to advance computational social science by investigating causal effects among societal events from observational data.
The causal effects among different types of societal events have not been extensively studied. 
In this work, we provide preliminary results on estimating the individual causal effects of one type of event on another and incorporate this causal information to improve the predictive power of event prediction models.
We hope to provide a way to understand human behavior from the societal and causal inference aspects and broaden the possibilities for future work on societal event studies.


\bibliographystyle{unsrt}  
\bibliography{references}

\end{document}